\documentclass[conference]{IEEEtran}
\usepackage{times}

\usepackage[numbers]{natbib}
\usepackage{multicol}
\usepackage[bookmarks=true]{hyperref}
\usepackage{amsmath}
\usepackage{graphicx}
\usepackage{hyperref}
\usepackage{xcolor}
\usepackage{amsfonts}       
\usepackage{algorithm}
\usepackage[noend]{algpseudocode}
\makeatletter
\newlength{\comment@width}
\usepackage{comment}
\usepackage{caption}
\usepackage{ragged2e}
\usepackage{subcaption}
\renewcommand{\Comment}[1]{%
  \sbox0{#1}
  \ifdim\wd0>\comment@width
    \setlength{\comment@width}{\wd0}%
  \fi
  \ifcsname comment@\arabic{algorithm}@width\endcsname
    \algorithmiccomment{\makebox[\csname comment@\thealgorithm @width\endcsname][l]{#1}}%
  \else
    \algorithmiccomment{#1}%
  \fi
}
\AtBeginEnvironment{algorithmic}{\setlength{\comment@width}{0pt}}
\AtEndEnvironment{algorithmic}{%
  \immediate\write\@auxout{%
    \string\algcommentwidth{\thealgorithm}{\the\comment@width}%
  }%
}
\newcommand{\algcommentwidth}[2]{%
  \global\@namedef{comment@#1@width}{#2}%
}
\makeatother

\usepackage{float}

\algnewcommand{\Inputs}[1]{%
  \State \textbf{Inputs:}
  \Statex \hspace*{\algorithmicindent}\parbox[t]{.8\linewidth}{\raggedright #1}
}
\algnewcommand{\Initialize}[1]{%
  \State \textbf{Initialize:}
  \Statex \hspace*{\algorithmicindent}\parbox[t]{.8\linewidth}{\raggedright #1}
}
\usepackage{caption}
\begin{document}





%


\title{Progressive Learning for Physics-informed Neural Motion Planning}
\author{Ruiqi Ni and Ahmed H. Qureshi
\\
Department of Computer Science, Purdue University
\\
\texttt{\{ni117,ahqureshi\}@purdue.edu}
}


\twocolumn[
{
\begin{@twocolumnfalse}
    \maketitle
    \vspace{-0.2in}
  \end{@twocolumnfalse}
\begin{figure}[H]
  \setlength{\hsize}{\textwidth}
    
  \centering
  \includegraphics[width=0.9\textwidth]{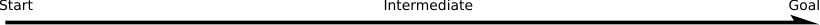}
  \\[0.1cm]
    \includegraphics[width=0.19\textwidth]{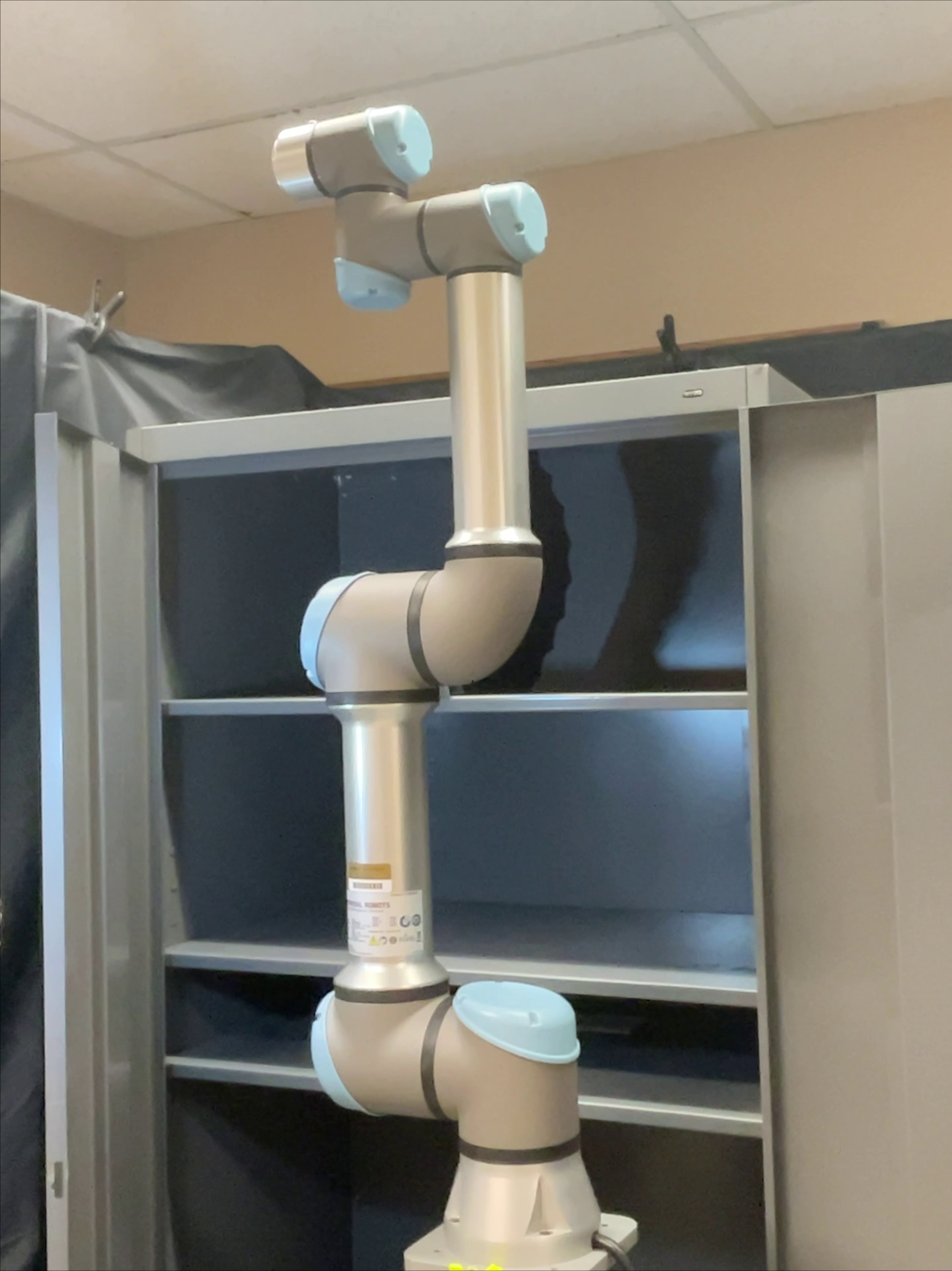}
    \includegraphics[width=0.19\textwidth]{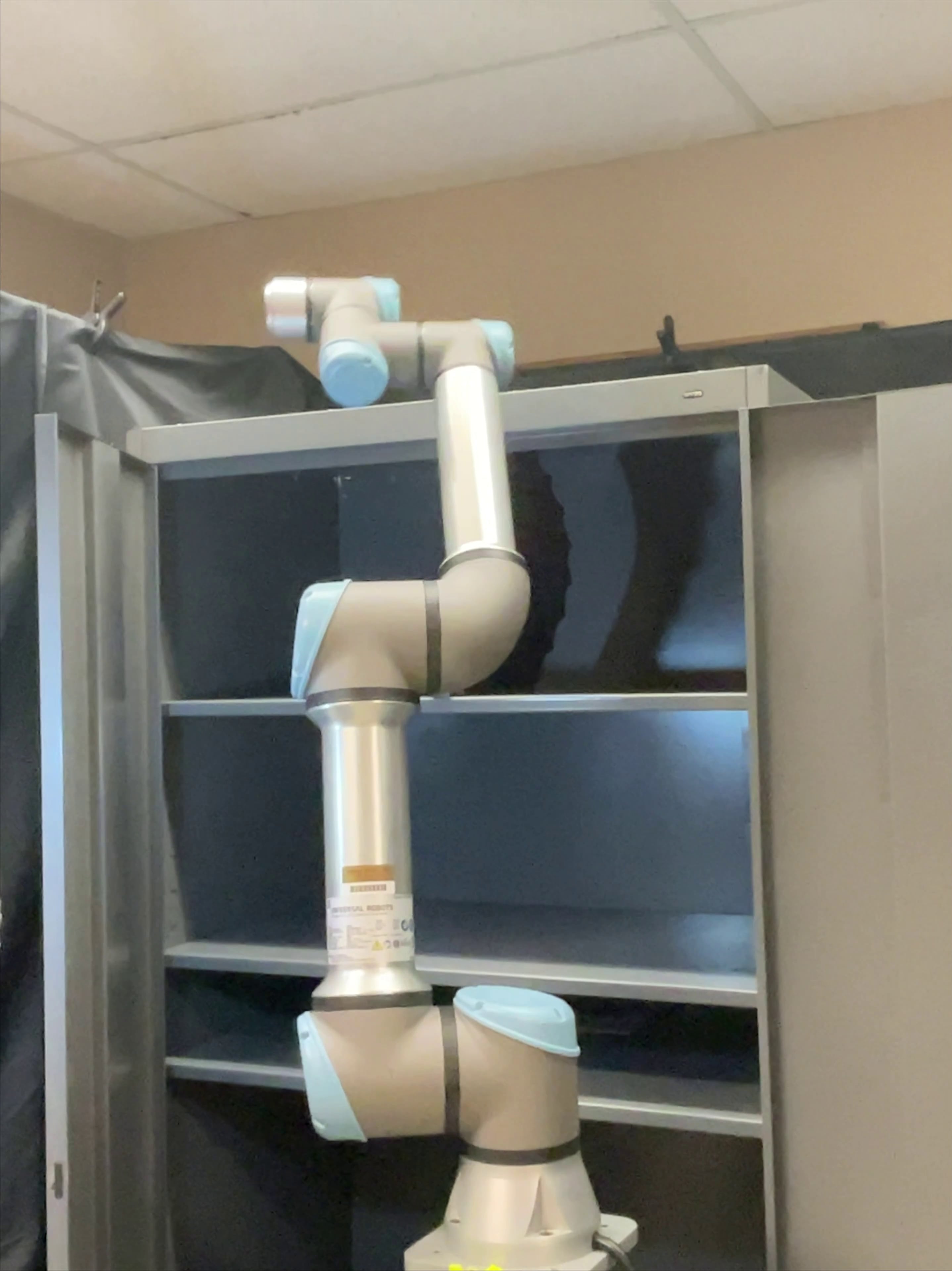}
    \includegraphics[width=0.19\textwidth]{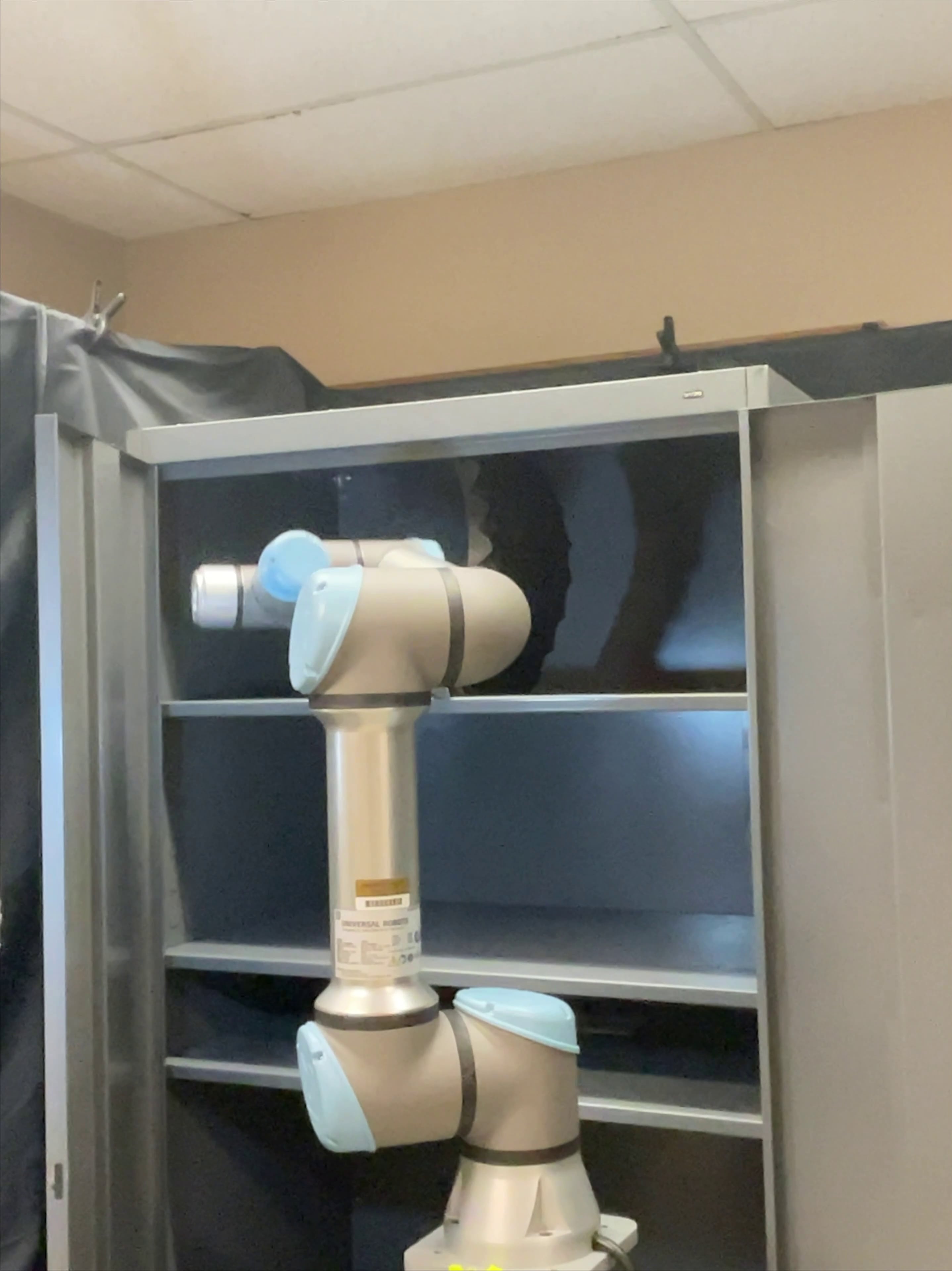}
    \includegraphics[width=0.19\textwidth]{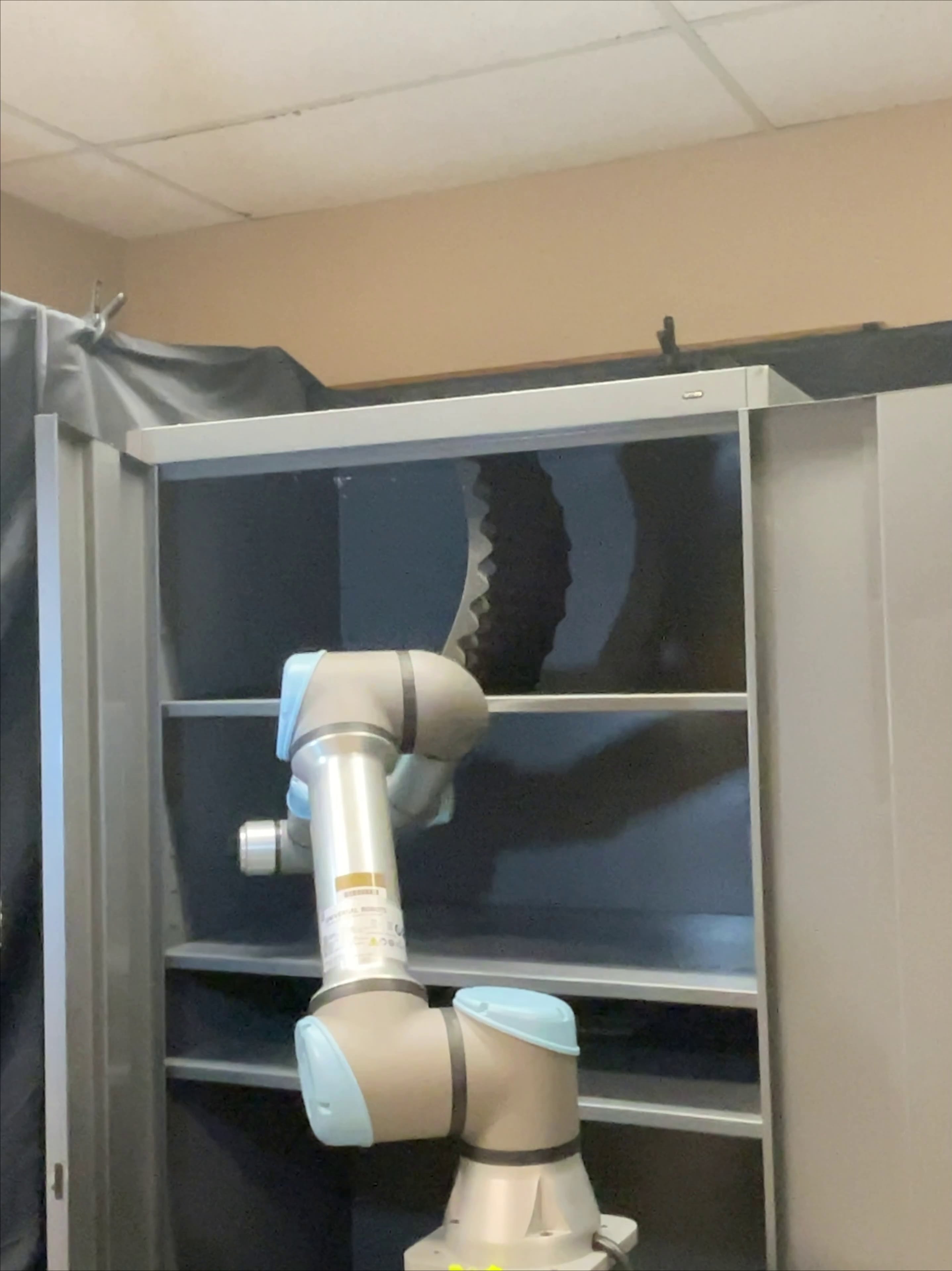}
    \includegraphics[width=0.19\textwidth]{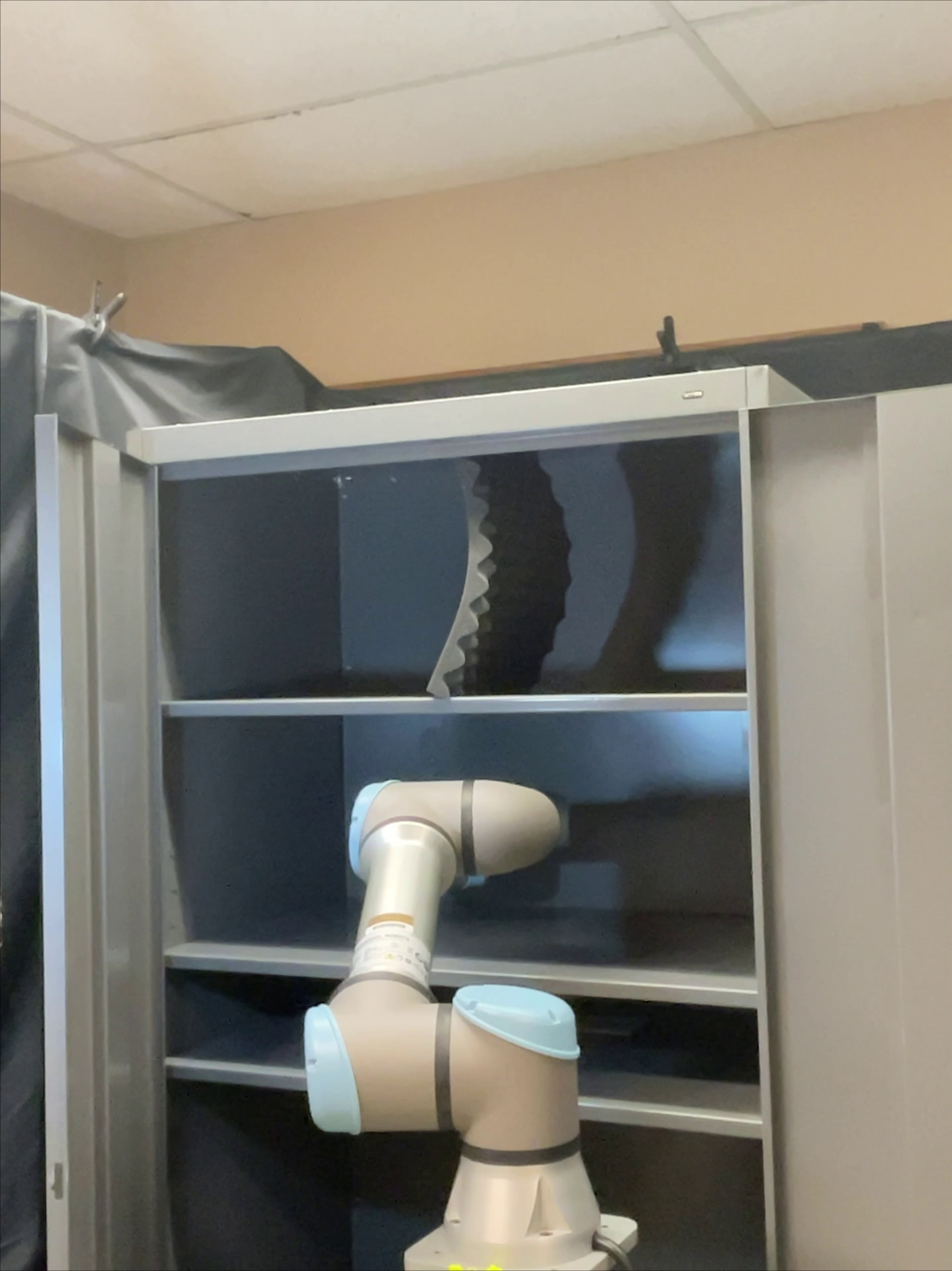} 
  \caption{\justifying Physics-informed neural motion planning of a 6-DOF robot manipulator in a real-world narrow passage environment. The images from left to right show the robot's motion sequence from its start to the desired goal configuration. In this case, the proposed approach took 0.05 seconds, whereas LazyPRM* took 2.79 seconds to find a path, making our method at least 50$\times$ faster than a traditional approach.}
  \label{fig:real_arm1}
  \vspace{-0.1in}
  \end{figure}}]

\begin{abstract}
Motion planning (MP) is one of the core robotics problems requiring fast methods for finding a collision-free robot motion path connecting the given start and goal states. Neural motion planners (NMPs) demonstrate fast computational speed in finding path solutions but require a huge amount of expert trajectories for learning, thus adding a significant training computational load. In contrast, recent advancements have also led to a physics-informed NMP approach that directly solves the Eikonal equation for motion planning and does not require expert demonstrations for learning. However, experiments show that the physics-informed NMP approach performs poorly in complex environments and lacks scalability in multiple scenarios and high-dimensional real robot settings. To overcome these limitations, this paper presents a novel and tractable Eikonal equation formulation and introduces a new progressive learning strategy to train neural networks without expert data in complex, cluttered, multiple high-dimensional robot motion planning scenarios. The results demonstrate that our method outperforms state-of-the-art traditional MP, data-driven NMP, and physics-informed NMP methods by a significant margin in terms of computational planning speed, path quality, and success rates. We also show that our approach scales to multiple complex, cluttered scenarios and the real robot set up in a narrow passage environment. The proposed method's videos and code implementations are available at \url{https://github.com/ruiqini/P-NTFields}.
\end{abstract}

\IEEEpeerreviewmaketitle

\section{Introduction}
Robots moving in their surrounding environment must find their feasible motion trajectory coordinating their actuators to move from their start configuration to goal configuration while satisfying all the constraints, such as collision avoidance. Various approaches exist, from classical methods \citep{karaman2011sampling,lavalle2001rapidly,gammell2014informed,janson2015fast} to learning-based neural motion planners (NMPs) \citep{qureshi2019motion,qureshi2020motion,ichter2018learning, qureshi2018deeply, kumar2019lego,chaplot2021differentiable}, that solve motion planning problems. However, the classical techniques suffer from the curse of dimensionality, i.e., they exhibit high computational times in finding a solution \citep{hauser2015lazy,gammell2014informed,qureshi2016potential}. In contrast, the NMPs demonstrate fast computational speeds at test time but require massive training data containing robot motion trajectories in the given environments \citep{ni2023ntfields}.  

Inspired by physics-informed deep learning models \citep{raissi2019physics, smith2020eikonet}, recent development has led to a physics-informed NMP called Neural Time Fields (NTFields) \citep{ni2023ntfields} that require no expert training trajectories and instead directly learn to solve the Eikonal equation for motion planning. Once trained, NTFields output the speed and time fields in the given environment for the desired start and goal configuration. Time fields' gradients are then followed to retrieve the feasible path solution for the underlying MP problem. Although NTFields find path solutions extremely fast and require no expert data, they struggle in complex environments and do not scale well to multiple scenarios and high-dimensional planning problems. These limitations are mainly due to the following two reasons. First, the Eikonal equation formulation has an extremely sharp feature solution around low-speed obstacles, making it difficult for the underlying deep-learning model to converge and perform well in complex scenarios. Second, training deep neural models to solve PDEs is inherently challenging and requires advanced learning strategies and an expressive PDE formulation with a smooth loss landscape. 
 
Therefore, this paper addresses the limitations of NTFields and proposes a new progressive learning method, which also requires no training trajectories and scales very well to complex scenarios, including high-dimensional, real-world robot manipulator planning problems. The main contributions of the paper are summarized as follows:

\begin{itemize}
    \item We highlight that the Eikonal equation formulation for motion planning in NTFields can converge to incorrect local minimums during training, resulting in relatively low performance and incapability to scale to multiple, complex environments.
    \item We introduce a novel progressive speed scheduling strategy that iteratively guides neural model training from a constant high speed to a very low speed around obstacles in the environment, preventing incorrect local minimums when training physics-informed NMPs in complex, cluttered environments.
    \item We propose using the viscosity term \citep{crandall1983viscosity} based on the Laplacian operator in the Eikonal equation formulation to transform its ill-posed, non-linear behavior into a semi-linear elliptic representation with a unique smooth solution around low-speed obstacles. Our novel formulation leads to physics-informed NMPs that are scalable to complex scenarios. 
    \item We compare our approach with a wide range of existing classical and learning-based methods. Our results show our proposed method leads to significantly better performance than prior methods, requires no training trajectories, and scales to multiple scenarios for motion planning.
    \item We also demonstrate our framework performance using a 6 degree-of-freedom (DOF) UR5e robot in solving real-world narrow passage motion planning problems, as shown in Fig. \ref{fig:real_arm1}.
\end{itemize}
  
\section{Related Work}
The pursuit for fast, efficient, and scalable motion planning methods began with complete \citep{lozano1979algorithm} and resolution-complete~\citep{khatib1986potential} techniques, which struggled in high-dimensional problems. A new class of sampling-based motion planning methods (SMPs)~\citep{kavraki1998analysis,bohlin2000path, kuffner2000rrt} emerged in early 2000, followed by their optimal variants~\citep{karaman2011sampling}, and have remained an industry-standard tool for almost over a decade. SMPs sample the robot configurations to build a graph in an obstacle-free C-space and then use Dijkstra-like algorithms \citep{dijkstra2022note} to retrieve paths connecting the given start and goal pairs. However, these methods also exhibit very low-computational speeds for finding path solutions. Therefore, several adaptive sampling approaches~\cite{qureshi2016potential, tahir2018potentially, gammell2014informed} have been proposed to bias the samples to the space containing the path solution to speed up the planning times. Although adaptive sampling methods are better than standard SMPs, they also struggle with the curse of dimensionality.

Recently, a new class of methods called Neural Motion Planners (NMPs)~\citep{qureshi2019motion,qureshi2020motion,ichter2018learning, qureshi2018deeply, kumar2019lego,chaplot2021differentiable} surfaced that find a path extremely fast at test time than traditional approaches and scale to high-dimensional problems with multi-DOF robot systems. However, the bottleneck to these methods is their need for many expert trajectories to train neural networks for motion planning. These expert trajectories often come from traditional planners such as SMPs, adding significant data generation computational load. In a similar vein, \citep{huh2021cost, li2022learning} perform supervised learning using data from conventional planners to learn the neural network-based cost-to-go (c2g) function or implicit environment functions (IEF). At test time, the gradients of the c2g function or IEF are followed to do the path planning. Another class of methods utilized Deep Reinforcement Learning to learn value functions for path planning~\citep{tamar2016value}. Those methods require many interactions with the environment for data generation and learning. Therefore, they are mostly demonstrated in toy problems, not real high-DOF robot settings. 

The most relevant work to our approach that solves the Eikonal equation and generates time fields for motion planning includes Fast Marching Method (FMM) \citep{sethian1996fast,valero2013path,treister2016fast}, and NTFields \citep{ni2023ntfields}. FMM is a classical approach that discretizes the robot C-space and uses wave propagation to find a solution to the Eikonal equation. Since FMM relies on discretization, it is computationally intractable in high-dimensional robot C-space. In contrast, the NTFields method is the first and most recent NMP that directly solves the Eikonal equation, does not require expert trajectories for training, and finds paths relatively faster than prior methods during testing. 

Although NTFields are shown to scale to high-DOF problems to some extent, they inherit limitations caused by the ill-posed nature of the Eikonal equation and the difficulty of training neural networks with physics equations. Therefore, NTFields could not scale to multiple environments and real-world settings and exhibited relatively lower success rates. We discuss these limitations in detail in the following sections and propose our new method that solves the Eikonal equation, requires no expert data, and finds solutions with very high success rates and low computation times. We also demonstrate the generalization of our approach to multiple environments and high-DOF real-world robot MP problems.

Aside from MP methods, it is also important to discuss parallel advances in PINNs and their relevance to our approach. PINNs \citep{raissi2019physics} provide a way for solving PDE by minimizing the PDE residual loss. Specifically, EikoNet \citep{smith2020eikonet} solves the Eikonal equation by PINN. However, their Eikonal equation formulation is not suitable for the MP tasks, as highlighted in \citep{ni2023ntfields}. Furthermore, progressively changing the reference signal has also been studied as the numerical continuation \citep{allgower2003introduction} in numerical methods. In PINNs, such phenomena are often coined as curriculum learning  \cite{krishnapriyan2021characterizing}. However, the curriculum learning for PINN focuses more on 1-dimensional PDE with simple boundary conditions, whereas we focus on real-world motion planning applications. Finally, the viscosity term has also been employed recently for signed distance fields (SDF) reconstruction \citep{lipman2021phase,pumarola2022visco}. However, the shortest distance solution of SDF is always a straight line, while MP requires a curved path to go around low-speed obstacles and induce complex collision avoidance constraints.

\section{Background}
This section formally presents the background to robot motion planning problems and their solutions through physics-informed NMPs. 
\subsection{Robot Motion Planning}
Let the robot's configuration and environment space be denoted as $\mathcal{Q}\subset \mathbb{R}^d$ and $\mathcal{X} \subset \mathbb{R}^m$, where $\{m,d\}\in \mathbb{N}$ represents their dimensionality. The obstacles in the environment, denoted as $\mathcal{X}_{obs} \subset \mathcal{X}$, form a formidable robot configuration space (c-space) defined as $\mathcal{Q}_{obs} \subset \mathcal{Q}$. Finally, the feasible space in the environment and c-space is represented as  $\mathcal{X}_{free}= \mathcal{X}\backslash \mathcal{X}_{obs}$ and $\mathcal{Q}_{free}= \mathcal{Q}\backslash \mathcal{Q}_{obs}$, respectively.  The objective of robot motion planning algorithms is to find a trajectory $\tau \subset \mathcal{Q}_{free}$ that connects the given robot start $q_s \in Q_{free}$ and goal $q_g \in Q_{free}$ configurations. Furthermore, additional constraints are sometimes imposed on the trajectory connecting the start and goal, such as having the shortest Euclidean distance or minimum travel time. The latter is often preferred as it allows imposing speed constraints near obstacles for robot and environment safety. However, planning under speed constraints is computationally expensive, and existing methods rely on path-smoothing techniques when safety is desired. 

\subsection{Physics-informed Motion Planning Framework}
Recent development led to a physics-informed motion planning framework called Neural Time Fields (NTFields) \citep{ni2023ntfields}, which provide a computationally-efficient and demonstration-free deep learning method for motion planning problems. It views motion planning problems as the solution to a PDE, specifically focusing on solving the Eikonal equation. The Eikonal equation, a first-order non-linear PDE, allows finding the shortest trajectory between start ($q_s$) and goal ($q_g$) under speed constraints by relating a predefined speed model $S(q)$ at configuration $q_g$ to the arrival time $T(q_s,q_g)$ from $q_s$ to $q_g$ as follows:
\begin{equation}
\frac{1}{S(q_g)} = \|\nabla_{q_g} T(q_s,q_g)\| 
\label{eikonal}
\end{equation}
The $\nabla_{q_g} T(q_s,q_g)$ is the partial derivative of the arrival time $T(q_s,q_g)$ function with respect to $q_g$. Therefore, finding a trajectory connecting the given start and goal requires solving the PDE under a predefined speed model and arrival time function. 
The arrival time function in NTFields is factorized as follows:
\begin{equation}
T(q_s,q_g)=\cfrac{\|q_s-q_g\|}{\tau(q_s,q_g)}   
\label{factorized}
\end{equation}
The $\tau(q_s,q_g)$ is the factorized time field which is the output of NTFields' deep neural network for the given $q_s$ and $q_g$. Given the arrival time function in Eq. \ref{factorized}, the Eikonal equation in Eq \ref{eikonal} expands to the following using the chain rule:
\begin{equation}
S(q_g) = \frac{\tau^2(q_s,q_g)}{\sqrt{
\begin{aligned}
\tau^2(q_s,q_g) &- 2\tau(q_s,q_g)  (q_g-q_s) \cdot \nabla_{q_g} \tau(q_s,q_g) 
\\&+ \|q_s-q_g\|^2  \|\nabla_{q_g} \tau(q_s,q_g)\|^2 
\end{aligned}
}}   
\label{eikonalloss}
\end{equation}

Since the neural network in NTfields outputs the factorized time field $\tau$, the corresponding predicted speed is computed using the above equation. Furthermore, the NTField framework determines the ground truth speed using a predefined speed function: 
\begin{equation}
S^*(q)=\cfrac{s_{const}}{d_{max}}\times\mathrm{clip}(\boldsymbol{\mathrm{d}}(\boldsymbol{\mathrm{p}}(q),\mathcal{X}_{obs}), d_{min}, d_{max})  
\label{speed}
\end{equation}
where $\boldsymbol{\mathrm{d}}(\cdot,\cdot)$ is the minimal distance between robot surface points $\boldsymbol{\mathrm{p}}(q)$ at configuration $q$ and the environment obstacles $\mathcal{X}_{obs}$. The $d_{min}$, and $d_{max}$ are minimum and maximum distance thresholds, and the $s_{const}$ is a predefined speed constant; we normalize $s_{const}=1$ to represent the maximum speed in the free space, and $s_{min}=s_{const}\times d_{min}/d_{max}$ represents the minimum speed in the obstacle space. Finally, the NTFields neural framework is trained end-to-end using the isotropic loss function \ref{ntf_trainloss} between predicted $S$ and ground truth $S^*$ speeds, i.e., 

\begin{equation}
    \begin{aligned}
   &|1-\sqrt{S^*(q_s)/S(q_s)}|+|1-\sqrt{S(q_s)/S^*(q_s)}|+\\
    &|1-\sqrt{S^*(q_g)/S(q_g)}|+|1-\sqrt{S(q_g)/S^*(q_g)}|
    \end{aligned}
    \label{ntf_trainloss}
\end{equation}

\section{Proposed Method}
Although NTFields demonstrate the ability for efficient motion planning without expert training data, it exhibits relatively low success rates in complex, cluttered environments, including high-dimensional problems, and does not scale to multiple scenarios. We observed that these limitations are mainly because of the ill-posed nature of the Eikonal equation and that the physics-informed loss landscapes are hard to optimize in general. To overcome these limitations, we introduce a new progressive learning algorithm comprising a novel viscosity-based Eikonal equation formulation and a progressive speed update strategy to train physics-informed NMPs in multiple, complex, high-dimensional scenarios. 

\subsection{Viscosity-based Eikonal Equation}
The Eikonal equation's exact solution has several problems that lead to neural network fitting issues. First, the solution is not differentiable at every point in space, which means a neural network cannot approximate the solution very well, especially for the sharp feature in low-speed environments. Second, the gradient $\nabla_{q_g} T(q_s,q_g)$ is not unique at these non-smooth points, which will also cause the neural network fitting issue because training is based on the supervision of the gradient $\nabla_{q_g} T(q_s,q_g)$.

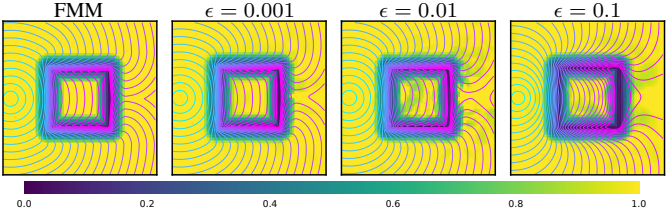
\begin{figure}[t]
    \centering
        \centering
        \includegraphics[width=0.49\textwidth,trim=0.3cm 0.29cm 0.2cm 0.2cm,clip]{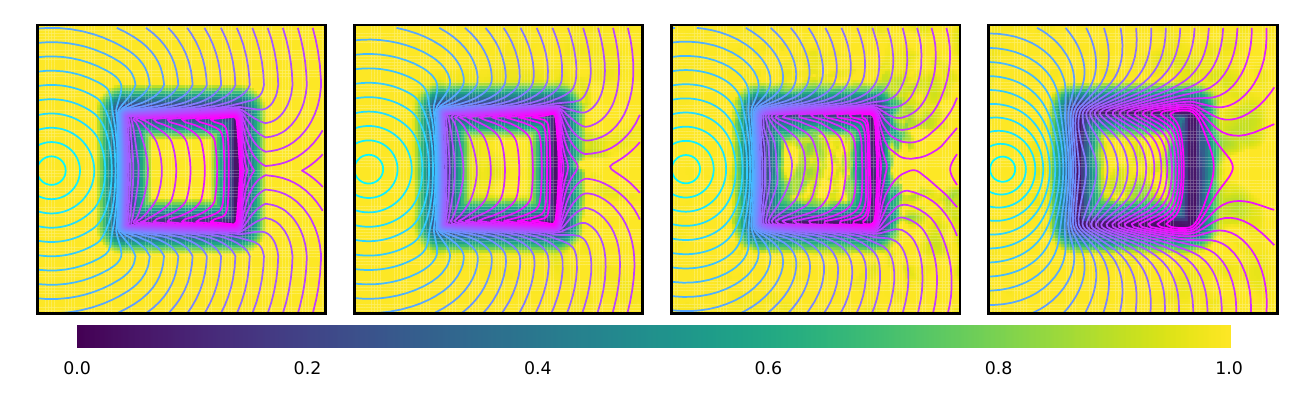}
    \put(-232,72){{\footnotesize FMM}}
    \put(-175,72){{\footnotesize $\epsilon=0.001$}}
    \put(-109,72){{\footnotesize $\epsilon=0.01$}}
    \put(-43,72){{\footnotesize $\epsilon=0.1$}}
    \caption{\justifying Effect of viscosity coefficient, $\epsilon$, on the correctness of time field results. It can be seen a large value of $\epsilon$ deviates from the solution given by the expert. The expert is FMM which finds a solution to the Eikonal equation. The colorbar shows the speed fields range from 0 to 1.} \vspace{-0.1in}
    \label{fig:visco}
\end{figure}

To fix these problems, we propose to use a viscosity term that can provide a differentiable and unique approximation of the Eikonal equation's solution. 
The viscosity term comes from the vanishing viscosity method \citep{crandall1983viscosity}. It adds the Laplacian $\Delta_{q_g} T(q_s,q_g)$ 
to the Eikonal equation, i.e., 
\begin{equation}
\frac{1}{S(q_g)} = \|\nabla_{q_g} T(q_s,q_g)\| + \epsilon \Delta_{q_g} T(q_s,q_g),
\label{vis}
\end{equation}
where $\epsilon \in \mathbb{R}$ is a scaling coefficient. The resulting system in Eq.~\ref{vis} is a semi-linear elliptic PDE with a smooth and unique solution. The expansion of Eq.~\ref{vis} using the chain rule and the $T$ described in Eq. \ref{viscoloss} becomes:  
\begin{equation}
S(q_g) = \frac{1}{\epsilon \Delta_{q_g} \tau(q_s,q_g)+\sqrt{
\begin{aligned}
 [\tau^2(q_s,q_g)   - 2\tau(q_s,q_g)  (q_g-q_s)
\\\cdot \nabla_{q_g} \tau(q_s,q_g)+\|q_s-q_g\|^2 
\\\times\|\nabla_{q_g} \tau(q_s,q_g)\|^2]/\tau^4(q_s,q_g) 
\end{aligned}
}}   
\label{viscoloss}
\end{equation}
Note that in the above equation, we use $\Delta_{q_g} \tau(q_s,q_g)$ instead of $\Delta_{q_g} T(q_s,q_g)$ for computational simplification, which also keeps the similar second-order derivative term. Furthermore, the value of $\epsilon$ affects the smoothness of the predicted time fields. In Fig \ref{fig:visco}, we compare fields with different values of $\epsilon$ to the ground truth field generated with the FMM approach. 
It can be seen that by varying the $\epsilon$, the correctness of results varies compared to the ground truth. In practice, when the coefficient  $\epsilon\rightarrow0$, the smooth and unique solution of Eq.~\ref{vis} will approach the exact solution of the Eikonal equation Eq.~\ref{eikonal}.

The above formulation resolves the ill-posed issue of the Eikonal equation. However, computing the Laplacian operator is computationally expensive as the existing deep learning libraries determine the hessian matrix of $\tau(q_s,q_g)$ and its trace to extract Laplacian. To mitigate the computational load, we employ the following two strategies. First, we only use the viscosity term during the training by setting $\epsilon$ to a constant value, and at planning time, we only compute $\nabla_{q_g} T(q_s,q_g)$ and $\nabla_{q_s} T(q_s,q_g)$. Second, we directly compute the diagonal of the hessian matrix using auto-differentiation following the strategy described in \citep{rall1981automatic}, resulting in similar complexity as standard gradient forward propagation, which is faster than complete hessian computation in standard deep learning libraries.

\subsection{Progressive speed scheduling}
This section introduces our progressive speed scheduling approach to train physics-informed motion planners in complex environments. The physics-based loss functions are generally challenging to optimize as they depend on the gradient of the underlying neural network. In physics-informed motion planners, the optimization becomes more difficult due to low-speed conditions near obstacles, often leading to an incorrect local minimum, i.e., despite small training loss, the neural model behaves as if low-speed obstacles do not exist in the environment. 
To circumvent the incorrect local minimums, we observe and leverage the following two properties of the Eikonal equation to progressively guide the NN training process and capture the low-speed obstacle space for collision avoidance.

\begin{figure}[t]
    \centering
        \centering
        \includegraphics[width=0.49\textwidth,trim=0.3cm 0.29cm 0.2cm 0.2cm,clip]{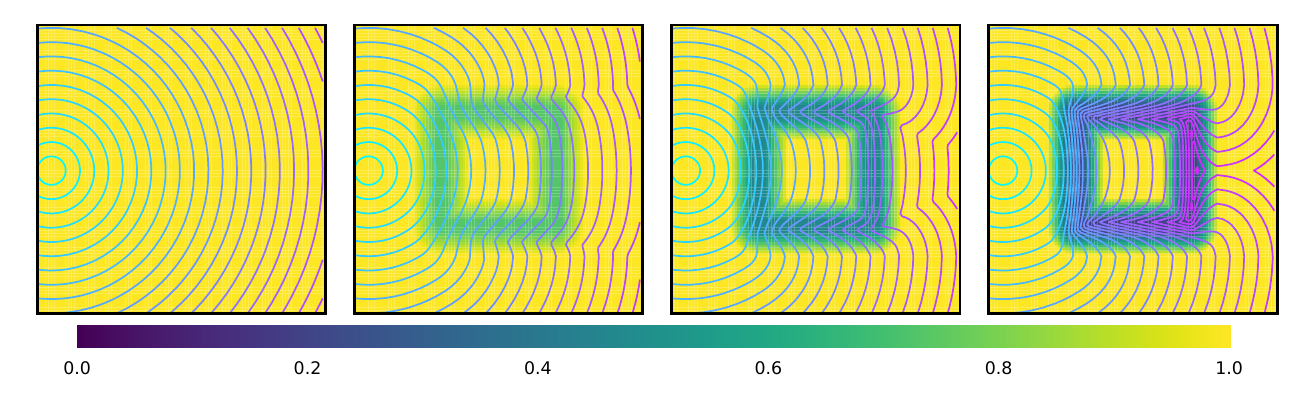}
    \put(-232,72){{\footnotesize $\alpha=0.0$}}
    \put(-172,72){{\footnotesize $\alpha=1/3$}}
    \put(-110,72){{\footnotesize $\alpha=2/3$}}
    \put(-44,72){{\footnotesize $\alpha=1.0$}}
    \put(-197,72){{{\color{black}{\footnotesize$\longrightarrow$}}}}
    \put(-133,72){{{\color{black}{\footnotesize$\longrightarrow$}}}}
    \put(-69,72){{{\color{black}{\footnotesize$\longrightarrow$}}}}
    \caption{\justifying Progressively decreasing the speed around obstacles using parameter $\alpha$ leads to continuous interpolation of speed and time fields in the given environment. The colorbar shows the speed fields range from 0 to 1.}\vspace{-0.1in}
    \label{fig:interp}
\end{figure}
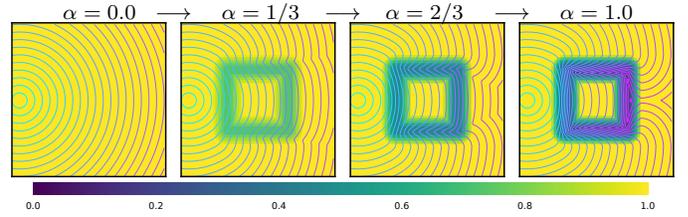

First, we notice the solution of the Eikonal equation (Eq. \ref{eikonal}), $T(q_s,q_g)$, in a constant max speed scene ($S(q)=1$) will become the distance between the given start and goal, which leads to trivial solution $\tau(q_s,q_g)=1$. Second, we find that the interpolation from the constant max-speed to the low speed around obstacles is continuous, and the solutions of the Eikonal equation along those interpolations are also continuous. 
Based on these observations, we propose a progressive speed alteration strategy that gradually scales down the speed from a constant max value to a low value around obstacles using a parameter $\alpha (t) \in [0,1]$, i.e.,
\begin{equation}
 S^*_{\alpha(t)}(q)=(1-\alpha(t))+ \alpha(t) S^*(q),  
\end{equation}
where $t \in \mathbb{N}$ represent the training epochs. 
Therefore, when $\alpha(t)=0$, the scene will have a constant max speed, and the Eikonal equation solution will be trivial. Furthermore, when $\alpha (t)=1$, the scene will have low speed around obstacles. We can also make $\alpha (t)>1$, such that the scene's minimal speed will become even lower than $s_{min}$. Fig \ref{fig:interp} shows the gradual progression of speed and time fields as $\alpha$ linearly scales from 0 to 1. It can be seen that the speed and time fields are changing continuously with $\alpha$ changing linearly. 

To train the physics-informed motion planner, we start with a low value of $\alpha(t)$ and let NN fit a constant speed trivial solution. 
Next, we progressively interpolate the field from constant max speed to low speed by gradually increasing the $\alpha(t)$ over the training epochs. The NN can easily fit the trivial solution. Then progressively decreasing obstacle speed $S^*(q)$ guides the network to learn the interpolating lower-speed fields. Furthermore, we also observe that the speed fields change linearly with $\alpha(t)$, but the resulting time fields change more aggressively. Thus, we also reduce the rate of change of $\alpha (t)$ as the training epochs increases. 

Lastly, since we gradually decrease the ground truth speed while training, the network parameters can change drastically and forget the previous learning. Several approaches exist to prevent such drastic changes, such as trust region optimization \citep{sorensen1982newton,schulman2015trust}. However, those approaches are often computationally expensive. We propose an alternative approach that bounds the ratio of loss $L$ at epoch $t$ and $t-1$ within a threshold $\eta$, i.e., $(L_t/L_{t-1})<\eta$. Whenever the loss ratio exceeds the imposed bound, we shuffle the training data so that the difficult samples causing high loss values are distributed across different training batches. That way, we observe the average loss does not diverge from the previous loss value and stays within our threshold. We also notice that without such a strategy, the physics-informed NN fails to converge and recover the underlying time field.

\subsection{Neural Architecture}
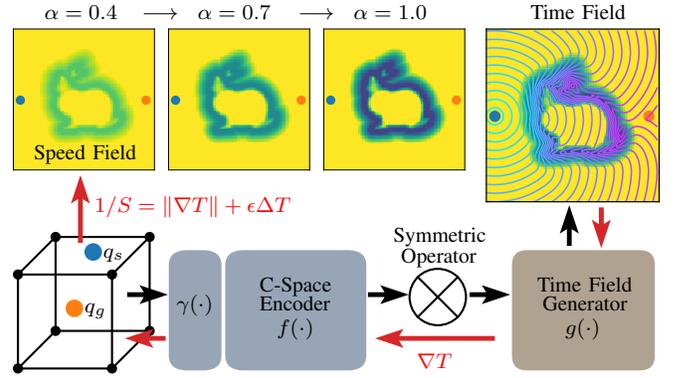
\begin{figure}[t]
    \centering
        \centering
        \includegraphics[width=0.49\textwidth,trim=0.0cm 0.0cm 0.0cm 0.0cm,clip]{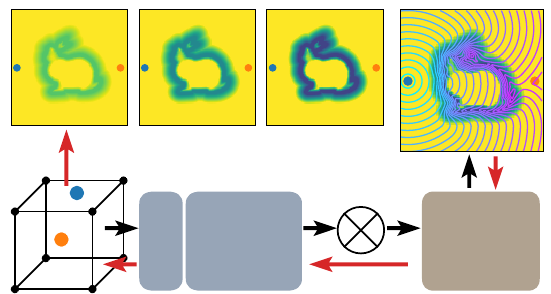}
    \put(-217,61){{{\color{red}{\footnotesize\begin{math}1/S=\|\nabla T\|+\epsilon\Delta T\end{math}}}}}
    \put(-236,134){{{\color{black}{\footnotesize$\alpha=0.4$}}}}
    \put(-199,134){{{\color{black}{\footnotesize$\longrightarrow$}}}}
    \put(-178,134){{{\color{black}{\footnotesize$\alpha=0.7$}}}}
    \put(-140,134){{{\color{black}{\footnotesize$\longrightarrow$}}}}
    \put(-119,134){{{\color{black}{\footnotesize$\alpha=1.0$}}}}
    \put(-95,2){{\color{red}{\footnotesize$\nabla T$}}}
    \put(-104,50){{\color{black}{\footnotesize Symmetric}}}
    \put(-101,42){{\color{black}{\footnotesize Operator}}}
    \put(-52,134){{\color{black}{\footnotesize Time Field}}}
    \put(-240,81){{\color{black}{\footnotesize Speed Field}}}
    \put(-213.7,44){{\footnotesize$q_s$}}
    \put(-220.7,23){{\footnotesize$q_g$}}
    \put(-186,24){{{\footnotesize\color{black}{\footnotesize$\gamma(\cdot)$}}}}
    \put(-154.5,31){{\footnotesize C-Space }}
    \put(-154.5,24){{\footnotesize  Encoder}}
    \put(-148,15){{{\footnotesize\color{black}$f(\cdot)$}}}
    \put(-49.5,31){{\footnotesize Time Field}}
    \put(-47.5,24){{\footnotesize Generator}}
    \put(-39,15){{{\footnotesize\color{black}$g(\cdot)$}}}

    \caption{ \justifying The neural architecture comprises the Fourier-based C-space Encoder, symmetric operator, and time-field generator. Three images on the top left show we progressively decrease the speed around a bunny-shaped obstacle to guide the neural network training. The image on the top right shows the final time field from start to goal generated by the trained model.}
    \label{fig:pipeline}
    \vspace{-0.2in}
\end{figure}

This section describes our neural framework, as shown in Fig. \ref{fig:pipeline}, for generating the speed and time fields for solving the robot motion planning problems. Our framework comprises the following modules.

\subsubsection{Fourier-based C-space Encoding}
Given the robot's initial ($q_s$) and target $(q_g)$ configurations, and the random environment latent code $\mathbf{b}\in \mathbf{R}^{d\times h}$, we compute the random Fourier feature $\gamma$ \citep{tancik2020fourier, rahimi2007random} for obtaining high-frequency robot configuration embeddings, i.e.,
\begin{equation}
\begin{aligned}
    &\gamma(q_s)=[\cos(2\pi \mathbf{b}^T q_s),\sin(2\pi \mathbf{b}^T q_s)]\\ &\gamma(q_g)=[\cos(2\pi \mathbf{b}^T q_g),\sin(2\pi \mathbf{b}^T q_g)]
    \end{aligned}
    \label{rff}
\end{equation}
The latent code $\mathbf{b}$ is of dimension $d\times h \in \mathbf{N}\times\mathbf{N}$ ($h$ is the hidden unit number) and represents the given environment. Although the latent code can be obtained in numerous ways, such as using auto-encoders to embed environment point clouds, we assign a fixed, unique random matrix to each environment for their representation. These features are further processed into a latent embedding by a C-space encoder $f(\cdot)$, which is a ResNet-style multi-layer perception \citep{he2016deep}.

\subsubsection{Non-linear Symmetric Operator}
To combine features $f(\gamma(q_s))$ and $f(\gamma(q_g))$, we use the non-linear symmetric operator $\bigotimes$ from NTFields method \citep{ni2023ntfields}. The operator $\bigotimes$ concatenates the $\max$ and $\min$ of two given features together, i.e., $f(\gamma(q_s))\bigotimes f(\gamma(q_g))=[\max(f(\gamma(q_s)),f(\gamma(q_g))), \min (f(\gamma(q_s)),f(\gamma(q_g)))]$. It is shown in \citep{ni2023ntfields} that the $\bigotimes$ operator is inspired by the Eikonal equation properties and leads to improved performance in predicting the speed and time fields. 
\subsubsection{Time Field Generator}
Our time field generator network $g$ is a ResNet-style multi-layer perceptron which takes the encoding $f(\gamma(q_s))\bigotimes f(\gamma(q_g))$ and outputs the factorized time field $\tau$, i.e.,
\begin{equation}
\tau(q_s,q_g)=g(f(\gamma(q_s))\bigotimes f(\gamma(q_g)))    
\end{equation}
Given the $\tau(q_s,q_g)$, we compute its gradient, $\nabla_{q_g} \tau(q_s,q_g)$, and Laplacian, $\Delta_{q_g} \tau(q_s,q_g)$, using auto-differentiation to determine the $S(q_s)$ and $S(q_g)$, as described in Eq. \ref{viscoloss}.
\subsection{Objective function}

The NTFields method uses L1-norm (Eq. \ref{ntf_trainloss}) to compute their isotropic loss function. The L1-norm is not smooth and imposes challenges converging to the optimal solution. Therefore, 
we propose a new, isotropic objective function, i.e.,
\begin{equation}
    \begin{aligned}
L(S^*_{\alpha}(q),S(q))=    &S^*_{\alpha}(q_s)/S(q_s)+S(q_s)/S^*_{\alpha}(q_s)+\\
    &S^*_{\alpha}(q_g)/S(q_g)+S(q_g)/S^*_{\alpha}(q_g)-4
    \end{aligned}
    \label{trainloss}
\end{equation}
Similar to the loss in \citep{ni2023ntfields}, our new loss is non-negative and has a minimum value of 0. However, unlike loss in \citep{ni2023ntfields}, the new loss is smooth and can easily reach the minimum value when $\alpha$ is small.

\subsection{Training pipeline}
Algorithm \ref{alg} outlines our training pipeline. The inputs to our procedure are described in Line 1, ranging from robot configurations dataset $D$ to neural architecture definition. As the training epoch, denoted by $i$, increases, the value of $\alpha$ is increased using predefined step sizes (Line 3). Consequently, the ground truth speed model is progressively modified at each training epoch using the parameter $\alpha$ (Line 4). Given the inputs and the ground truth $S^*_\alpha$, we begin the batch training by forming batches from dataset $D$. The dataset $D$ contains the valid robot start $(q_s)$ and goal $(q_g)$ configurations. For each start and goal pair in batch $B_j$, we predict the factorized time $\tau(q_s,q_g)$ and associated speeds, $S(q_s)$ and $S(q_g)$, using Equation \ref{viscoloss} (Line 8-9). Next, we compute the loss between predicted and ground truth speeds (Line 10) and aggregate the resulting average batch losses to compute the total dataset loss (Line 12). To prevent neural parameters $\theta$ from diverging, we reshuffle the samples between batches whenever $L_i/L_{i-1}>\eta$, where $\eta$ is a predefined threshold. The reshuffling is performed to distribute the difficult samples and prevent large gradient-based parameter updates at line 11. 

\begin{algorithm}[ht]
\caption{\label{alg:train} Progressive Learning}

\small{\begin{algorithmic}[1]
\Inputs{ 
$D$\Comment{Robot configurations dataset}\\
$S^*(q)$ \Comment{Groud truth speed model}\\
$\tau_{\theta}(\cdot)$ \Comment{Neural network model}\\
$\theta$ \Comment{Model parameters}\\
$\alpha, step$ \Comment{Progressive learning parameters}\\
\vspace{5pt}
}

\For{$i=1,\cdots$} \Comment{Training epoches}
\State $\alpha = \alpha + step$
\State $S^*_{\alpha}(q) = (1-\alpha)+\alpha S^*(q)$ \Comment{Compute progressive speed}

\State $L_{i} = 0$ \Comment{Begin batch training}
\For{batches $j=0, \cdots$} \Comment{Batch $B_j$ from $\mathcal{D}$}
    \State $\forall ({q_s},{q_s})\in B_j$ \Comment{Training data in one batch}
    \State $\tau_{\theta}({q_s},{q_g})$ \Comment{Predict factorized time}
    \State $S_{\theta}({q_s}),\; S_{\theta}({q_s})$ \Comment{Predict speed by Eq. \ref{viscoloss}}
    \State $l_j=L(S^*_{\alpha}(q),S_{\theta}(q))$ \Comment{Compute loss by Eq. \ref{trainloss}}
    \State $\theta=\theta-\nabla_{\theta}l_j$\Comment{Update model parameters}
    \State $L_{i} =L_{i}+l_j$
\EndFor
\If{$L_{i}/L_{i-1}>\eta$} \Comment{Check loss threshold}
      \State Reload $i$-th epoch model parameters $\theta$
      \State Reshuffle samples among batches
      \State Repeat batch training steps 5-12
\EndIf
\EndFor
\end{algorithmic}}
\label{alg}
\end{algorithm}

\subsection{Planning pipeline}
Once trained, we use the execution pipeline similar to the NTFields method. First, we predict $\tau(q_s,q_g)$ for the given start $q_s$, goal $q_g$, and latent environment code $\mathbf{b}$, as described in Eq. \ref{rff}. Next, the factorized time, $\tau$, parameterizes  Eq. \ref{factorized} and \ref{eikonalloss} for computing time $T(q_s,q_g)$ and speed fields $S(q_s), S(q_g)$, respectively. Note that, we use Eq. \ref{eikonalloss} instead of Eq. \ref{viscoloss} for efficient speed computations, i.e., without Laplacian $(\epsilon \rightarrow 0)$. Finally, the path solution is determined in a bidirectional manner by iteratively updating the start and goal configurations as follows,
\begin{equation}
    \begin{aligned}
    q_s &\gets q_s-\beta S^2(q_s)\nabla_{q_s} T(q_s,q_g) \\
    q_g &\gets q_g-\beta S^2(q_g)\nabla_{q_g} T(q_s,q_g)
    \end{aligned}
    \label{plan}
\end{equation}
The parameter $\beta \in \mathbb{R}$ is a predefined step size. Furthermore, at each planning iteration, the start and goal configurations are updated using gradients to march toward each other until $\|q_s-q_g\|<d_g$, where $d_g \in \mathbb{R}$. 

\subsection{Implementation Details:}
This section provides the implementation details including the data generation process and hyperparameters. 
\subsubsection{Data generation:} Our data generation process is the same as NTFields, i.e., we only need the valid start and goal configuration pairs across different environments for training. We present three training and testing scenarios: eight cluttered 3D (C3D) environments, two Gibson environments, and two narrow passage 6-DOF robot manipulator environments. For C3D and Gibson, we use 8$\times$0.5M and 2$\times$1M configuration pairs, respectively; for the manipulator, we use 2$\times$1M pairs. Furthermore, like NTFields, our data generation process is swift and takes less than two minutes to gather.
\subsubsection{Hyperparameters}

Our method hyperparameters include the following. For training, we set $\alpha=0.5$ for the first $1000$ epochs, then increase $\alpha$ with $step=1/4000$ for each epoch; when the epoch is greater than 4000, we reduce $step=1/8000$, until $\alpha>=1.05$ to get the final result. We set $\eta=1.5$ and use AdamW \citep{loshchilov2017decoupled} as the optimizer with the learning rate as $10^{-3}$ and the weight decay as $0.1$. For testing, we set $\beta = 0.03$ and $d_g=0.06$ for 3D environments, $\beta = 0.02$ and $d_g=0.04$ for 6-DOF manipulator.
Regarding neural network parameters, our method can make a smaller neural network to solve the Eikonal equation well: we reduce the hidden unit number of NTFields from 256 to 128. We also reduce the network ResNet-style block number of NTFields from 10 to 5. Finally, our network size is reduced from NTFields 50MB to 7MB. In that way, our method performs more efficiently in planning.

\section{EVALUATION}
In this section, we evaluate our method through the following experiments. First, we perform the ablation analysis demonstrating the effectiveness of our new Eikonal formulation and the progressive speed scheduling strategy. Second, we perform a comparative analysis to evaluate the performance of our method against a variety of state-of-the-art baselines. For this analysis, we consider three environment setups: (1) cluttered 3D (C3D) environment contains eight scenarios, each with 10 obstacles randomly placed in the 3D space. (2) Gibson environment in which we picked two scenarios from the Gibson dataset. (3) 6-DOF UR5e robot manipulator planning in two complex cabinet environments with narrow passages. For these scenarios, we present evaluations in both simulation and real-world.
We perform all experiments on a computing system with 3090 RTX GPU, Core i7 CPU, and 128GB RAM. The baseline methods and the evaluation metrics for the comparative analysis are summarized as follows:

\textbf{Baselines:}
\begin{itemize}
    \item \textbf{FMM:} A wave propagation method \citep{sethian1996fast} that discretizes the given C-space and solves the Eikonal equation to compute the arrival time for path planning.
    \item \textbf{RRT*:} A single-query SMP that constructs optimal trees and returns a path connecting the given start and goal. We set the 10 seconds time limit in which if RRT* retrieves a path, it is considered successful.
    \item \textbf{RRT-Connect:} A bidirectional trees method that grows two trees from the start and the goal until they meet. We further process the generated paths via smoothing to get shorter path lengths in the same homotopy class. Similar to RRT*, we set the 10 seconds time limit in which if RRT-Connect retrieves a path, it is considered successful.
    \item \textbf{Lazy-PRM*:} A multi-query SMP that randomly samples a given environment and constructs a connected graph. Then for any new start and goal pair, the graph is queried through the nearest neighbor search to extract the path solution. The theme of lazy methods is to minimize the number of collision checks, thus resulting in shorter planning times. Like RRT*, we also set the 10 seconds time limit for this planner to find a path solution.
    \item \textbf{IEF3D:} We use the  modified IEF3D shown in NTFields paper. It learns implicit environment functions for path planning using an encoder-decoder structure. The encoder is a PointNet module that takes the environment point cloud and outputs its latent embedding. The decoder is a ResNet-style MLP that takes the environment embedding, the start and goal configuration, and outputs the time field. IEF3D is trained via supervised learning and requires expert demonstration data. We randomly sampled start and goal pairs in C-space space and used FMM to find their Eikonal equation solution. The resulting data is used to train IEF3D via supervised learning. Furthermore, IEF3D can scale to different environments.
    \item \textbf{NTFields:} As described earlier, it directly solves the Eikonal equation and does not require expert training data. However, since NTFields do not scale to multiple environments, we train individual NTField models for each presented scenario. For instance, in the C3D environment, we have eight scenarios, so we had to train eight NTField models for testing.  
\end{itemize}

\textbf{Evaluation Metrics:}
\begin{itemize}
    \item \textbf{Time:} The planning time shows the computational time in seconds for the planner to find a valid collision-free path connecting the given start and goal.
    \item \textbf{Length:} The path length indicates the sum of Euclidean distance between path waypoints.
    \item \textbf{Safe margin:} The safety margin indicates the closest distance to obstacles along the path.
    \item \textbf{SR:}  Finally, the success rate (SR) shows the percentage of collision-free paths found by a planner in a given test dataset.
\end{itemize}

\begin{figure}[t]
    \centering
        \centering
        \includegraphics[width=0.49\textwidth,trim=0.3cm 0.29cm 0.2cm 0.2cm,clip]{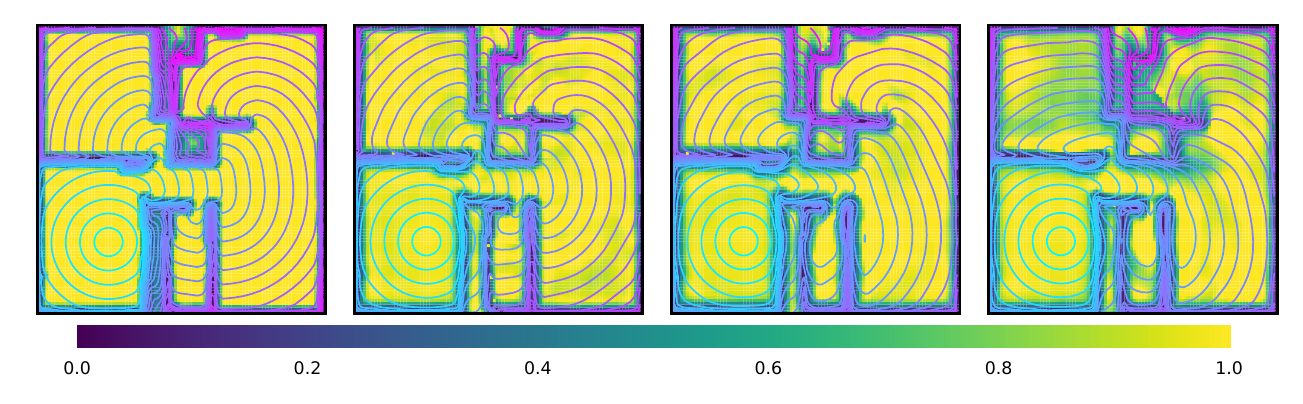}
    \put(-232,72){{\footnotesize FMM}}
    \put(-167,72){{\footnotesize Ours}}
    \put(-122,72){{\footnotesize w/o scheduling}}
    \put(-52,72){{\footnotesize w/o viscosity}}
    \caption{\justifying We compare our method with and without progressive speed scheduling or the viscosity term against FMM on a Gibson environment for time field generation. Our method recovers the correct result. The color bar shows the speed fields range from 0 to 1.}
    \label{fig:compare}
    \vspace{-0.2in}
\end{figure}
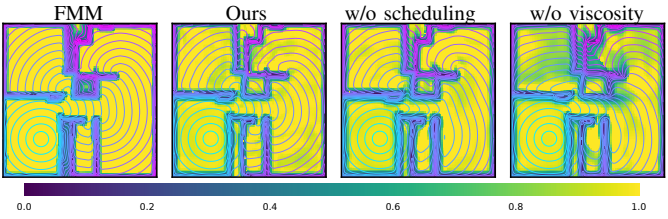

\begin{figure}[t]
\centering
        \centering
\begin{subfigure}[b]{2\linewidth}
        \includegraphics[width=0.24\textwidth, trim=14.3cm 0.3cm 12.2cm 0.2cm,clip]{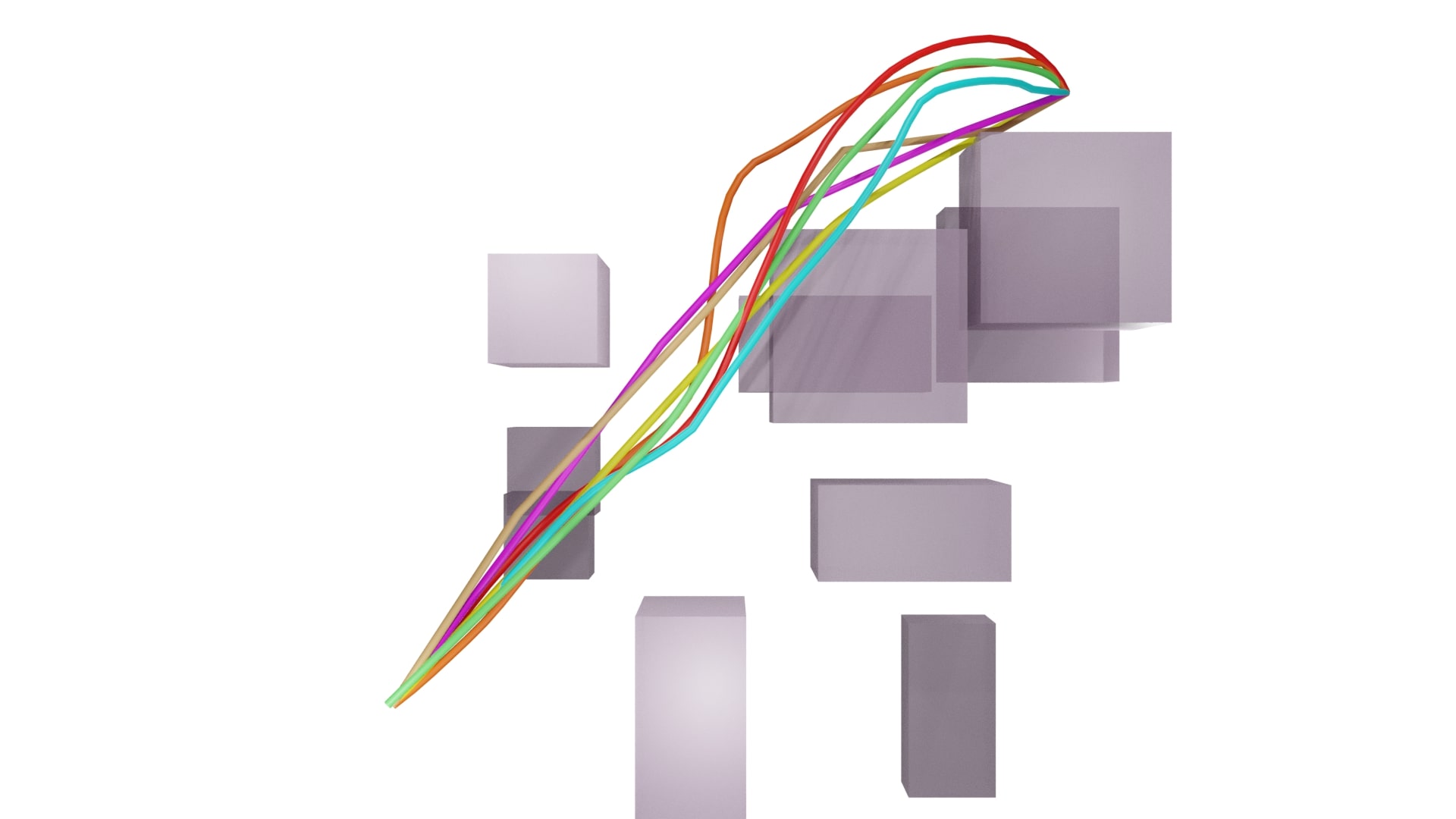}
        \includegraphics[width=0.24\textwidth,trim=15.3cm 4.2cm 10.2cm 2.2cm,clip]{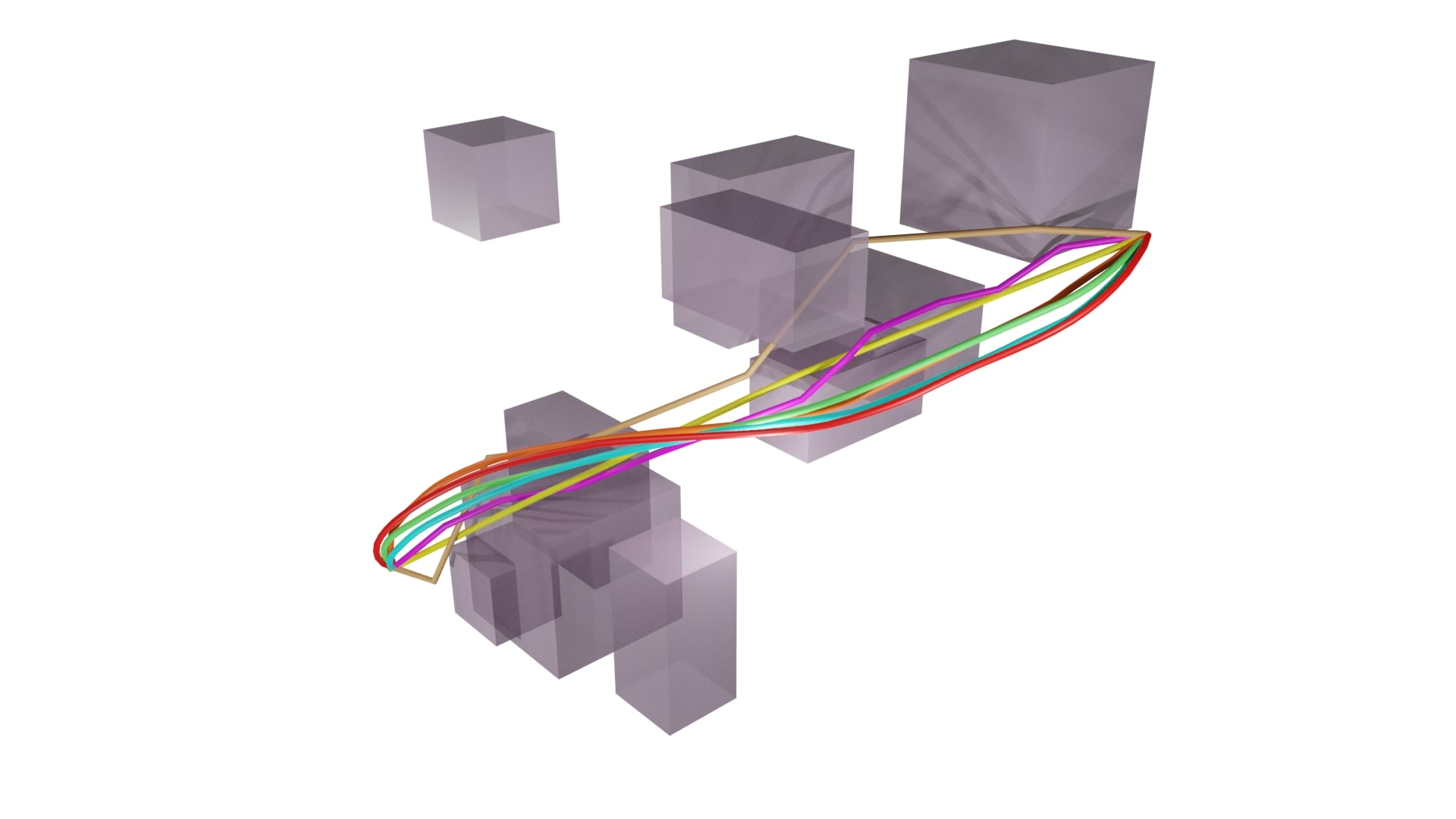}

    \vspace{0.1in}
\end{subfigure}
\begin{subtable}[b]{1\linewidth}
\footnotesize
\setlength{\tabcolsep}{5pt}
\begin{tabular}{ccccc}
\hline
C3D & time (sec)   &  length & {safe margin} & sr(\%)\\ 
\hline
Ours  & $0.02 \pm 0.00$ & $15.20 \pm 15.62$ & {$1.94 \pm 1.07$} & 99.8 \\
        NTFields  & $0.05 \pm 0.00$ & $15.47 \pm 15.74$ & {$2.09 \pm 1.06$} & 99.9 \\
        IEF3D  & $0.06 \pm 0.00$ & $15.21 \pm 15.59$ & {$1.89 \pm 1.11$} & 99.5 \\
        FMM   & $0.62 \pm 0.01$  & $15.03 \pm 14.08$ & {$1.96 \pm 1.00$} & 100\\
        RRT*   & $2.08 \pm 0.00$  & $14.49 \pm 13.90$ & {$1.31 \pm 1.63$} &99.8\\
       { LazyPRM*}   & {$1.38 \pm 8.16$}  & {$14.35 \pm 11.60$} & {$1.20 \pm 1.83$} & 99.5\\
       { RRT-Connect}   & {$0.11 \pm 0.02$}  & {$14.67 \pm 13.81$} & {$1.22 \pm 1.82$} & 100\\
\hline
\end{tabular}
\label{c3d}
\end{subtable}
\caption{\justifying Comparison in C3D environments. The figures show six paths generated by our method (orange), NTFields (red), IEF3D (cyan), FMM (green), RRT* (brown), LazyPRM* (pink), and RRT-Connect (yellow). The boxes are the obstacles. The statistical results on 8$\times$1000 different starts and goals for eight C3D environments.}
\label{fig:c3d}
\vspace{-0.25in}
\end{figure}

\subsection{Abalation Analysis}
In this section, we analyze our progressive learning framework by evaluating the role of progressive speed scheduling and the viscosity term in recovering time fields. Fig. \ref{fig:compare} shows the speed and time fields of our method and its ablations in comparison to FMM. The speed field is shown in the color scale, where color changing from yellow to blue indicates speed changing from maximum to minimum, respectively. The contours represent the time field from a start point. From Fig. \ref{fig:compare} contours, it can be seen that our approach gets a similar result as FMM, whereas without progressive speed scheduling or viscosity term, the same neural model cannot recover the correct result. Quantitatively, the loss of our method is very small, i.e., 0.015 units, whereas without speed scheduling, it increases to 0.03 units, and without viscosity, it further increases to 0.15 units. We also observed that NTFields could recover similar fields as our method but require a larger neural network size (50MB) than ours (7MB) and do not scale to multiple scenarios. Hence, the ablation analysis validates that our progressive learning aids in efficiently recovering and learning near-optimal time and speed fields in complex environments. 
\begin{figure}[b]
\centering
\begin{subfigure}[b]{2\linewidth}
    
        \includegraphics[width=0.21\textwidth, trim=15.3cm 0.3cm 16.2cm 0.2cm,clip]{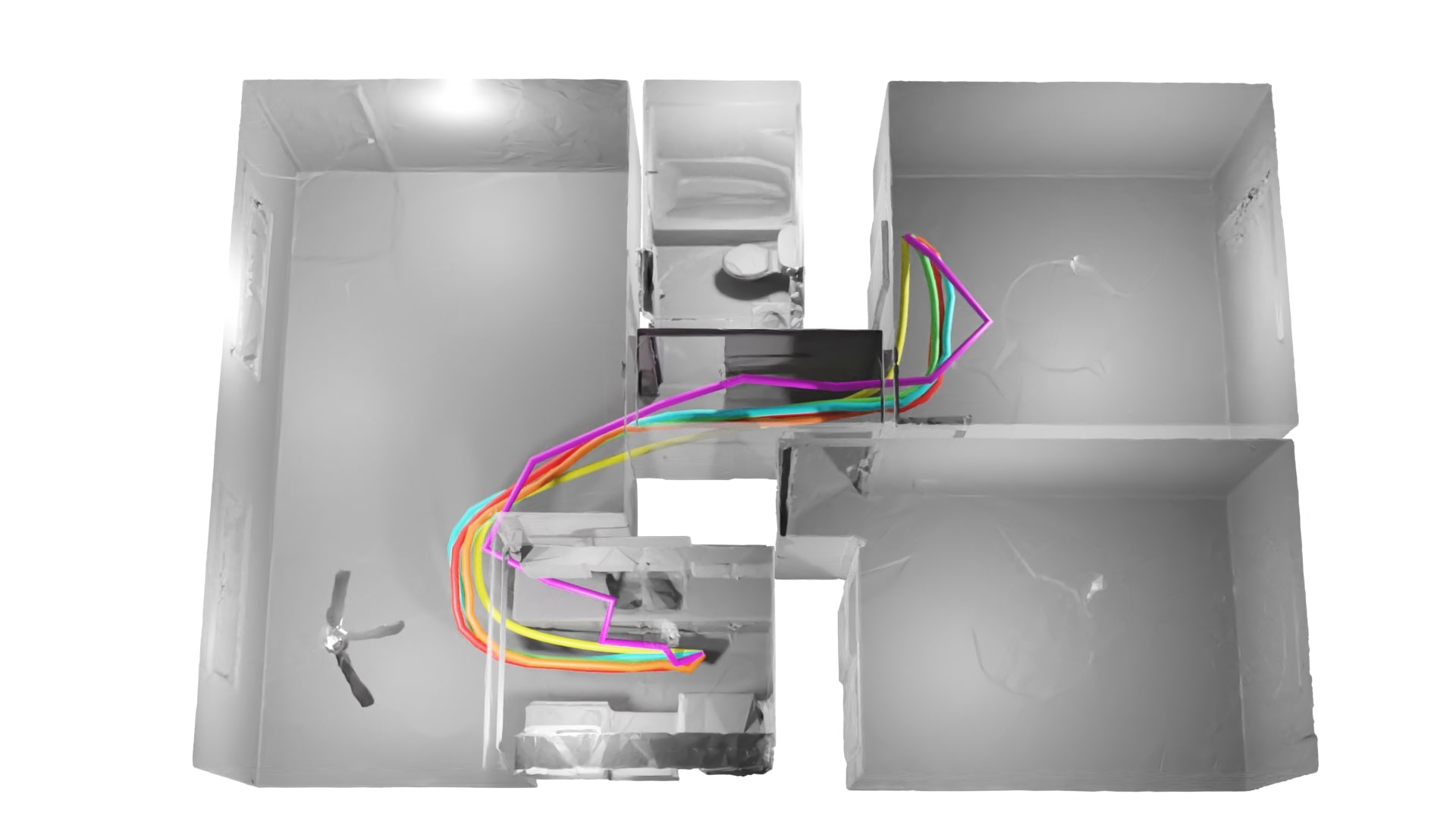}
        \includegraphics[width=0.27\textwidth,trim=6.3cm 1.3cm 16.2cm 1.2cm,clip]{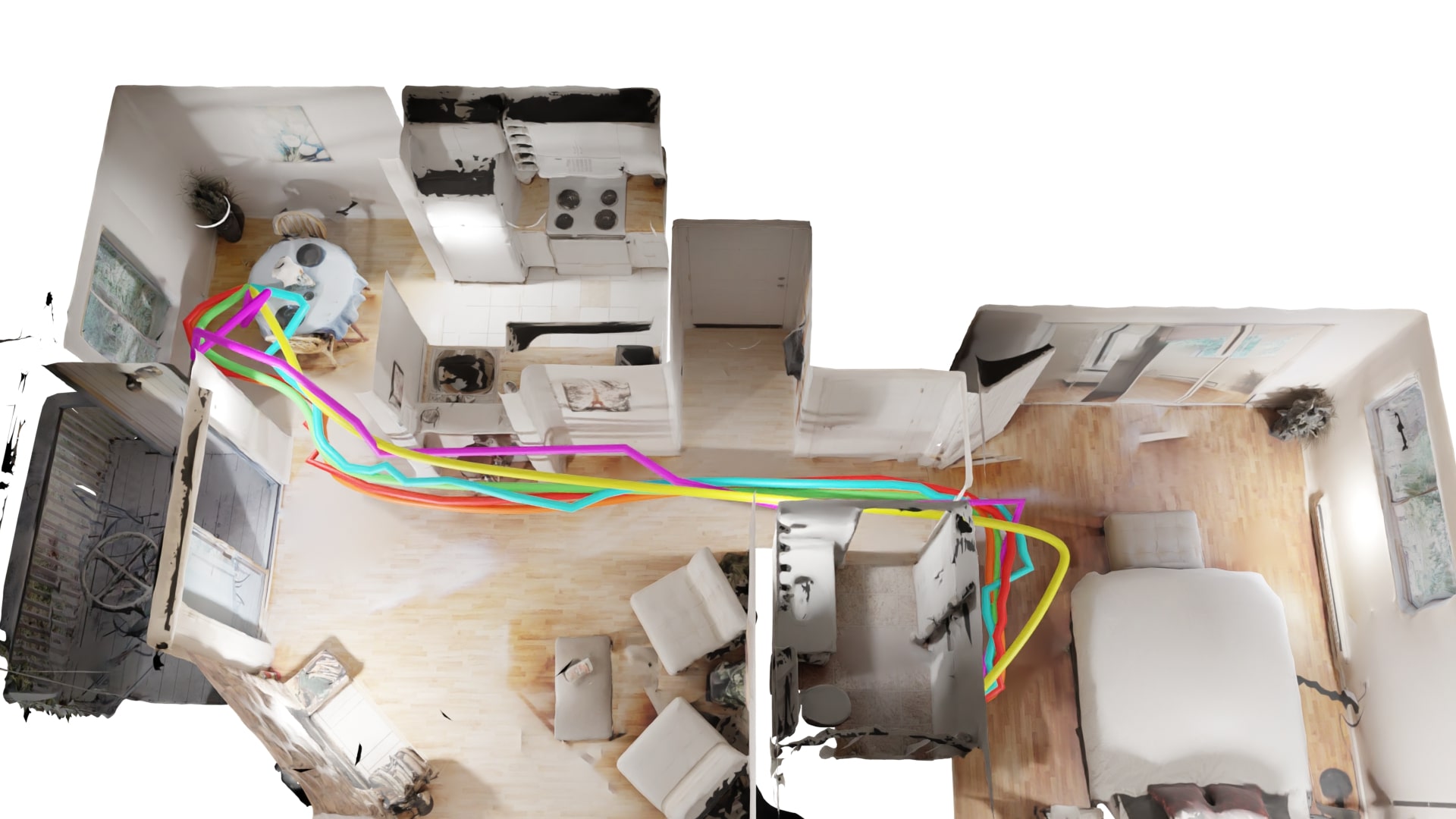}
    
    
    \vspace{0.1in}
\end{subfigure}

\begin{subtable}[h]{1\linewidth}
\footnotesize
\setlength{\tabcolsep}{5pt}
\begin{tabular}{ccccc}
\hline
Gibson & time (sec)   &  length & {safe margin} & sr(\%)\\ 
\hline
Ours  & $0.01 \pm 0.00$ & $11.68 \pm 29.69$ & {$0.88 \pm 0.16$} & 98.3 \\
       NTFields  & $0.03 \pm 0.00$ & $11.31 \pm 54.23$ & {$0.94 \pm 0.16$} & 91.6 \\
        IEF3D  & $0.05 \pm 0.00$ & $11.47 \pm 27.69$ & {$0.87 \pm 0.18$} & 92.5 \\
        FMM   & $0.69 \pm 0.00$  & $11.21 \pm 24.79$ & {$0.93 \pm 0.13$} & 97.4\\
        RRT*   & $3.17 \pm 0.00$  & $10.36 \pm 26.28$ & {$0.57 \pm 0.29$} &89.8\\
       { LazyPRM*}   & {$2.63 \pm 25.09$}  & {$9.94 \pm 16.27$} & {$0.53 \pm 0.35$} & 92.9\\
       {RRT-Connect}   & {$0.44 \pm 0.28$}  & {$11.95 \pm 32.88$} & {$0.56 \pm 0.34$} & 100\\
\hline
\end{tabular}

\label{gibson}
\end{subtable}

\caption{\justifying Comparison in two Gibson environments. The figures show six paths generated by our method (orange), NTFields (red), IEF3D (cyan), FMM (green), LazyPRM* (pink), and RRT-Connect (yellow). The statistical results on 2$\times$500 different starts and goals for two Gibson environments.}
\label{fig:gibson}
\end{figure}

\subsection{Comparison Analysis}
This section presents a comparative analysis of our method and other baselines on C3D, Gibson, and 6-DOF manipulator environments. Note that our method and IEF3D generalize to multiple environments, whereas for NTFields, we trained a separate neural model for each scenario.

\begin{figure*}
\centering
\includegraphics[width=0.8\textwidth]{fig/arr1.pdf}
  \\[0.1cm]
  \includegraphics[width=0.01\textwidth,trim=0.0cm -0.75cm 0.0cm 0.0cm,clip]{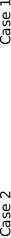}
  \begin{subfigure}[b]{0.97\textwidth}
         \centering
         \includegraphics[width=0.19\textwidth]{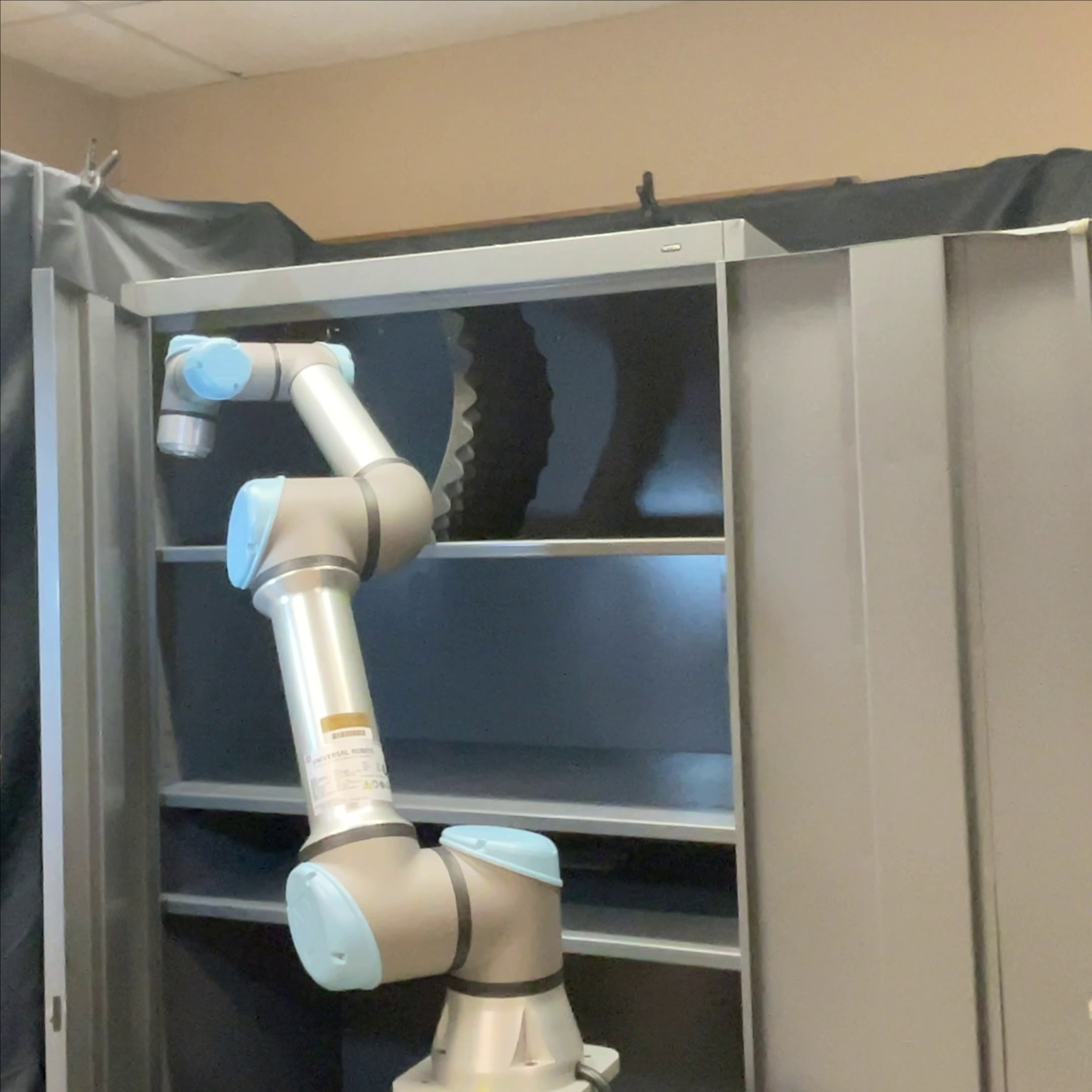}
    \includegraphics[width=0.19\textwidth]{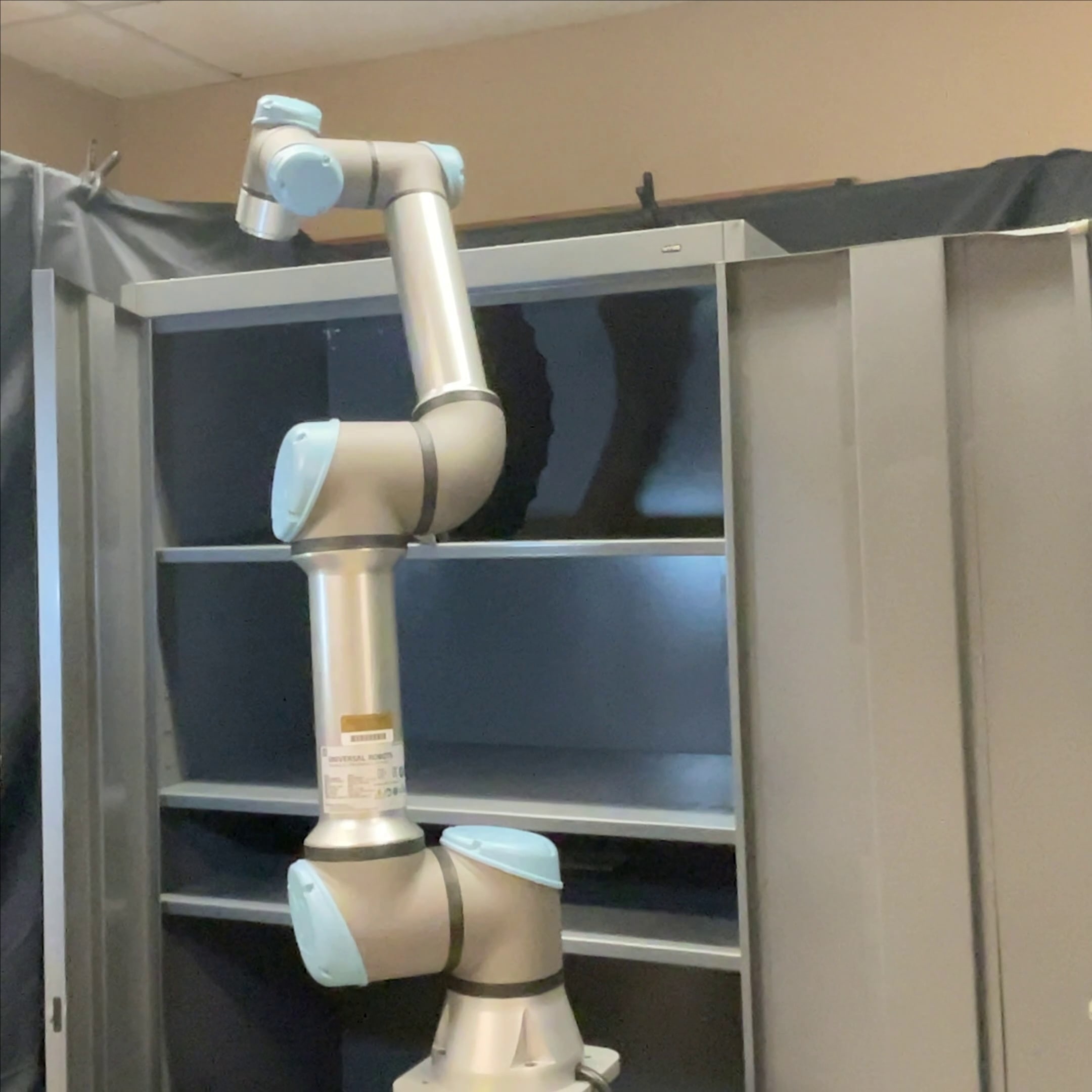}
    \includegraphics[width=0.19\textwidth]{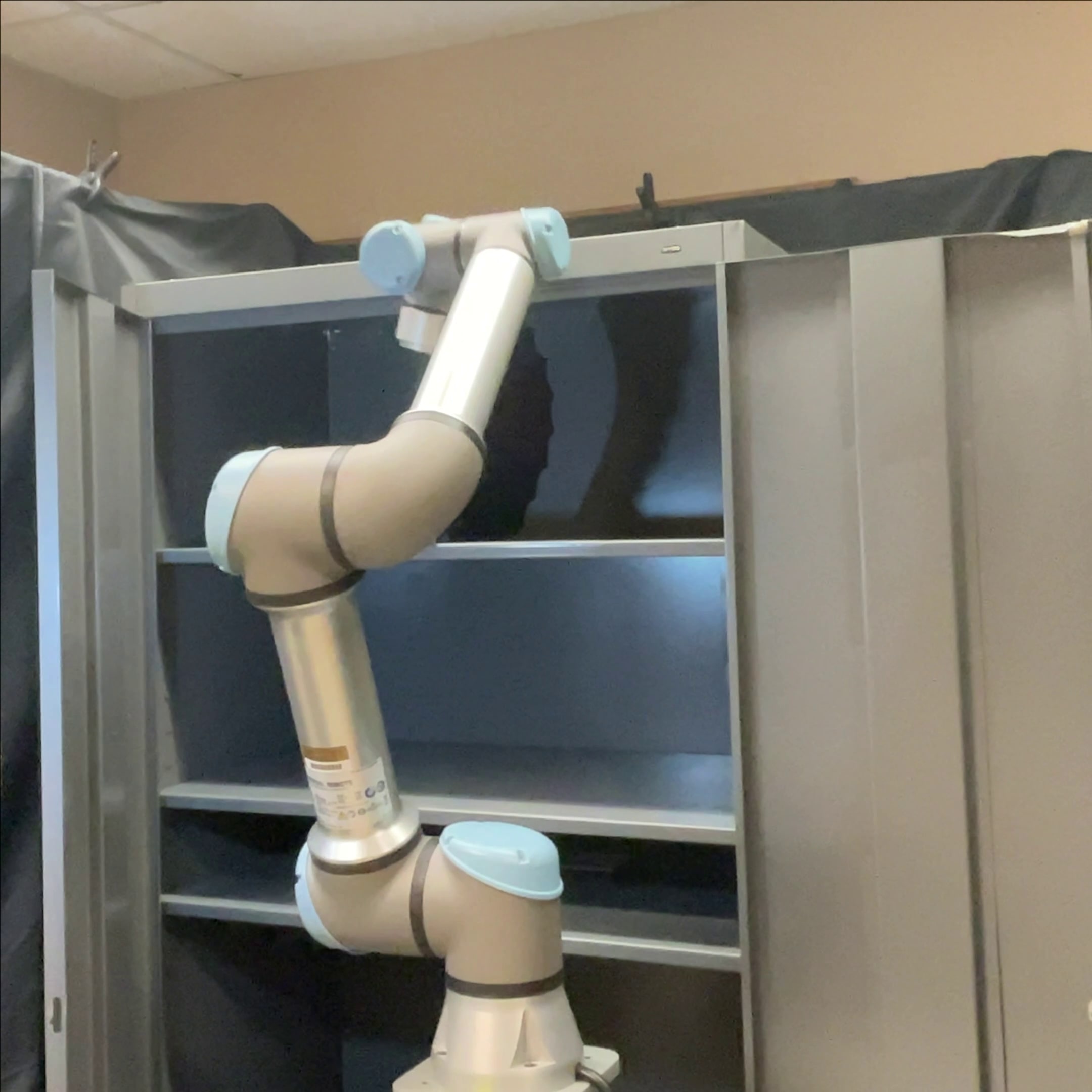}
    \includegraphics[width=0.19\textwidth]{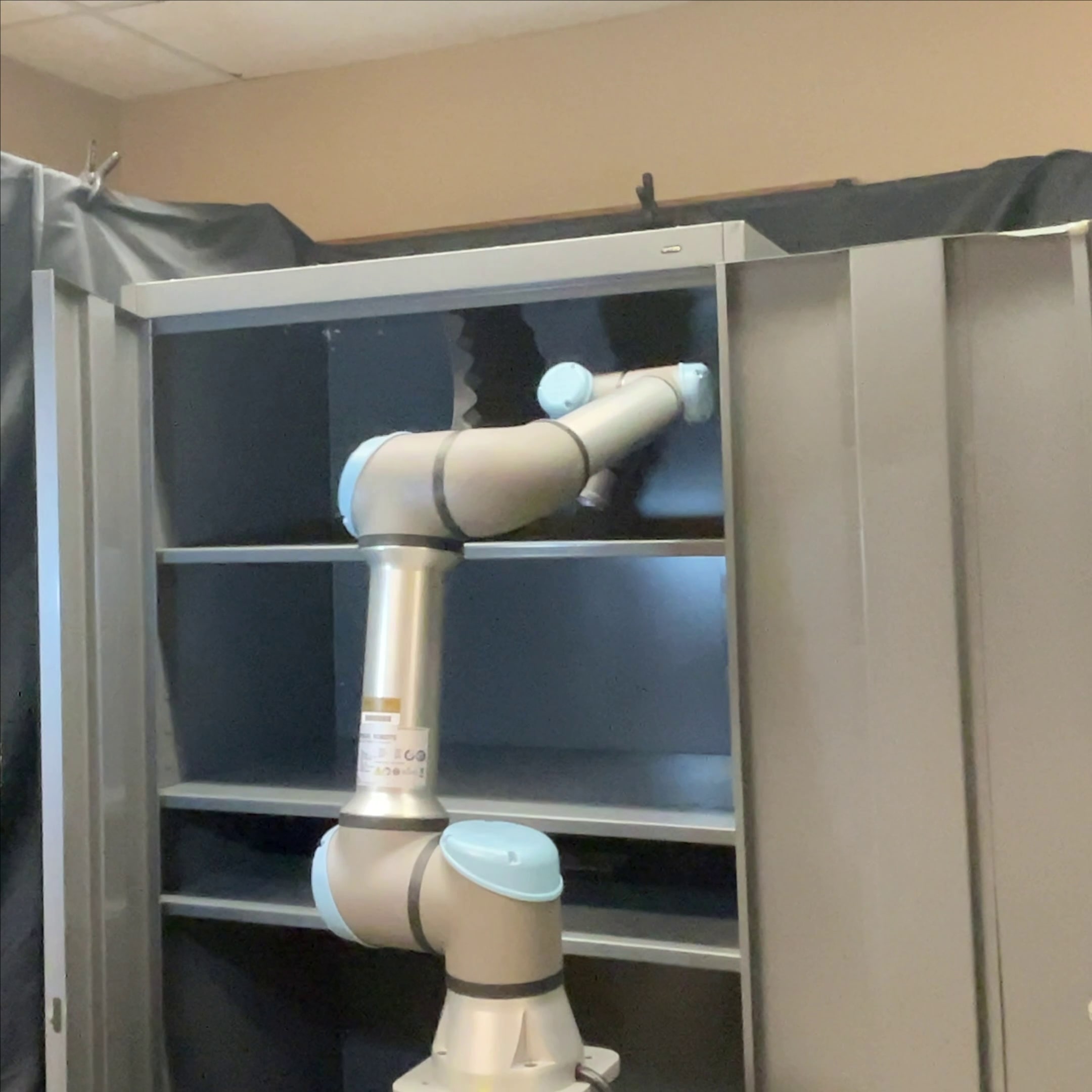}
    \includegraphics[width=0.19\textwidth]{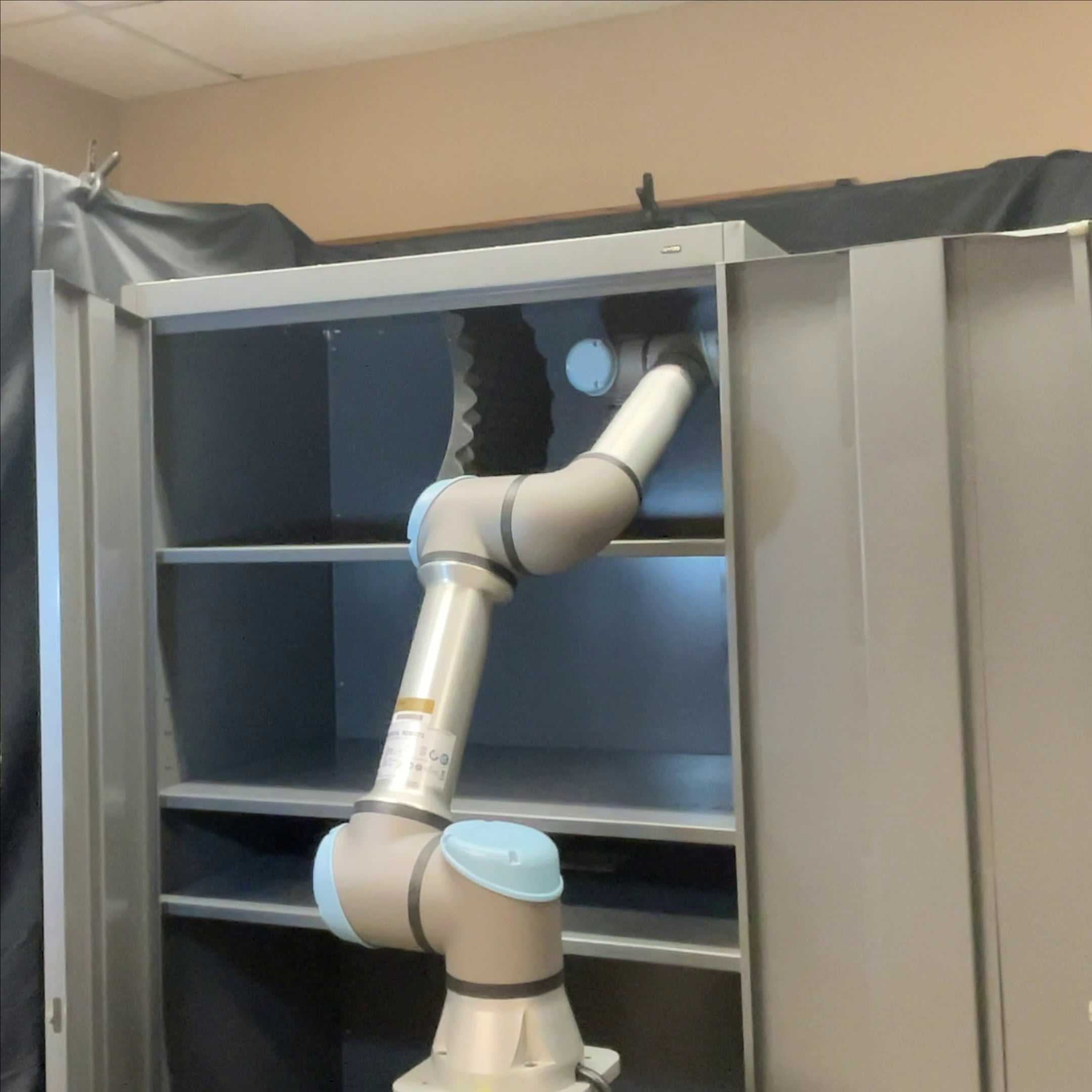}
    \\[0.1cm]
    \includegraphics[width=0.19\textwidth]{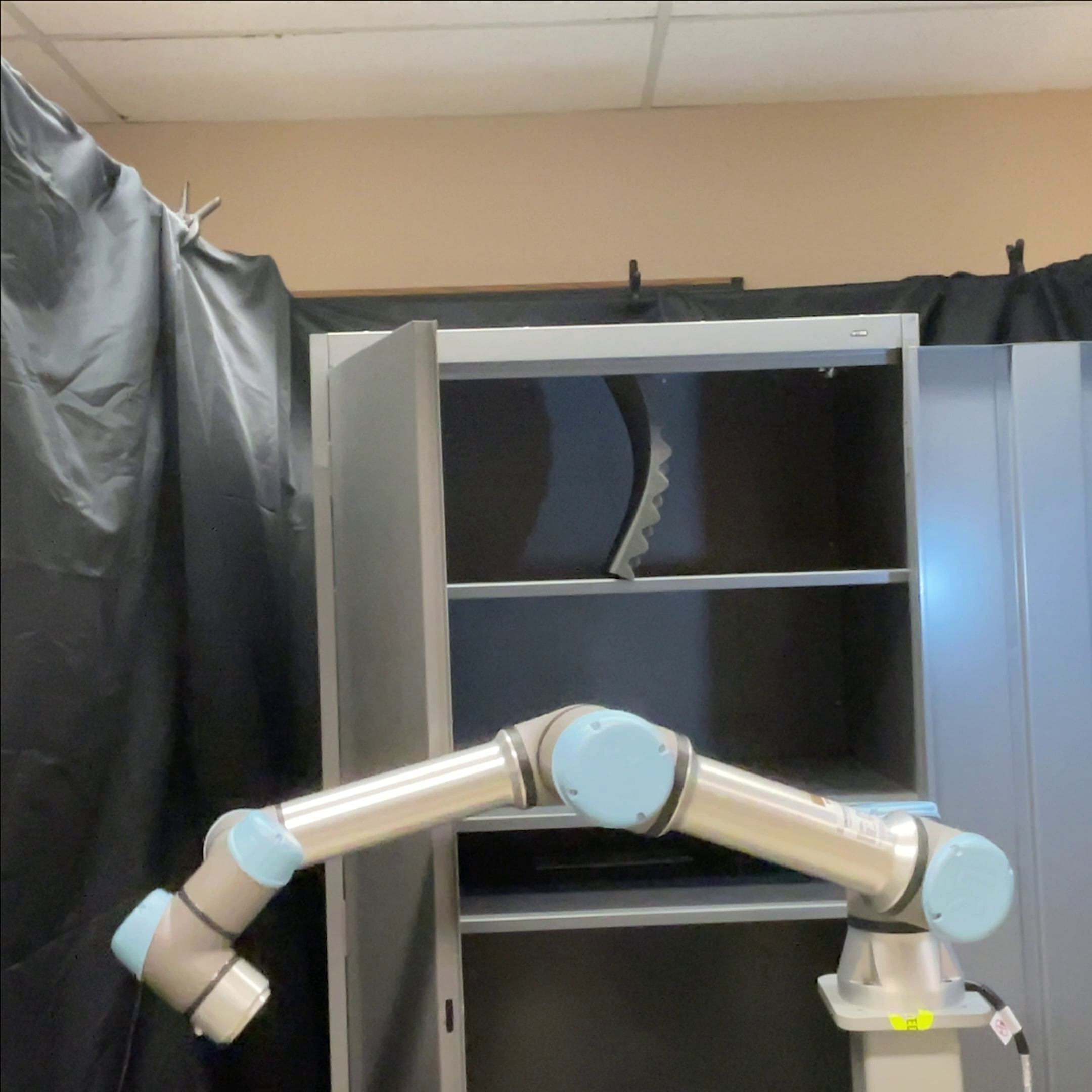}
    \includegraphics[width=0.19\textwidth]{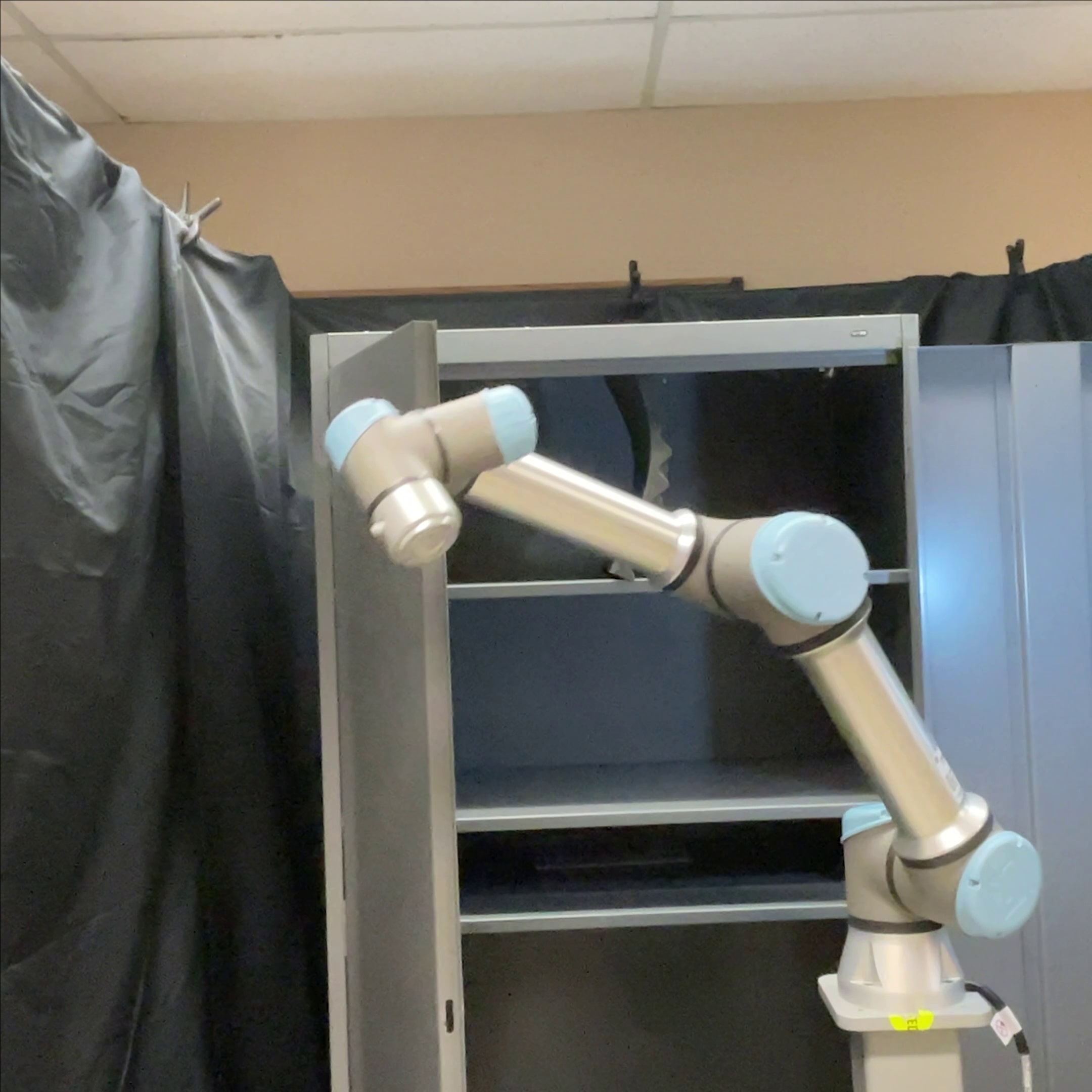}
    \includegraphics[width=0.19\textwidth]{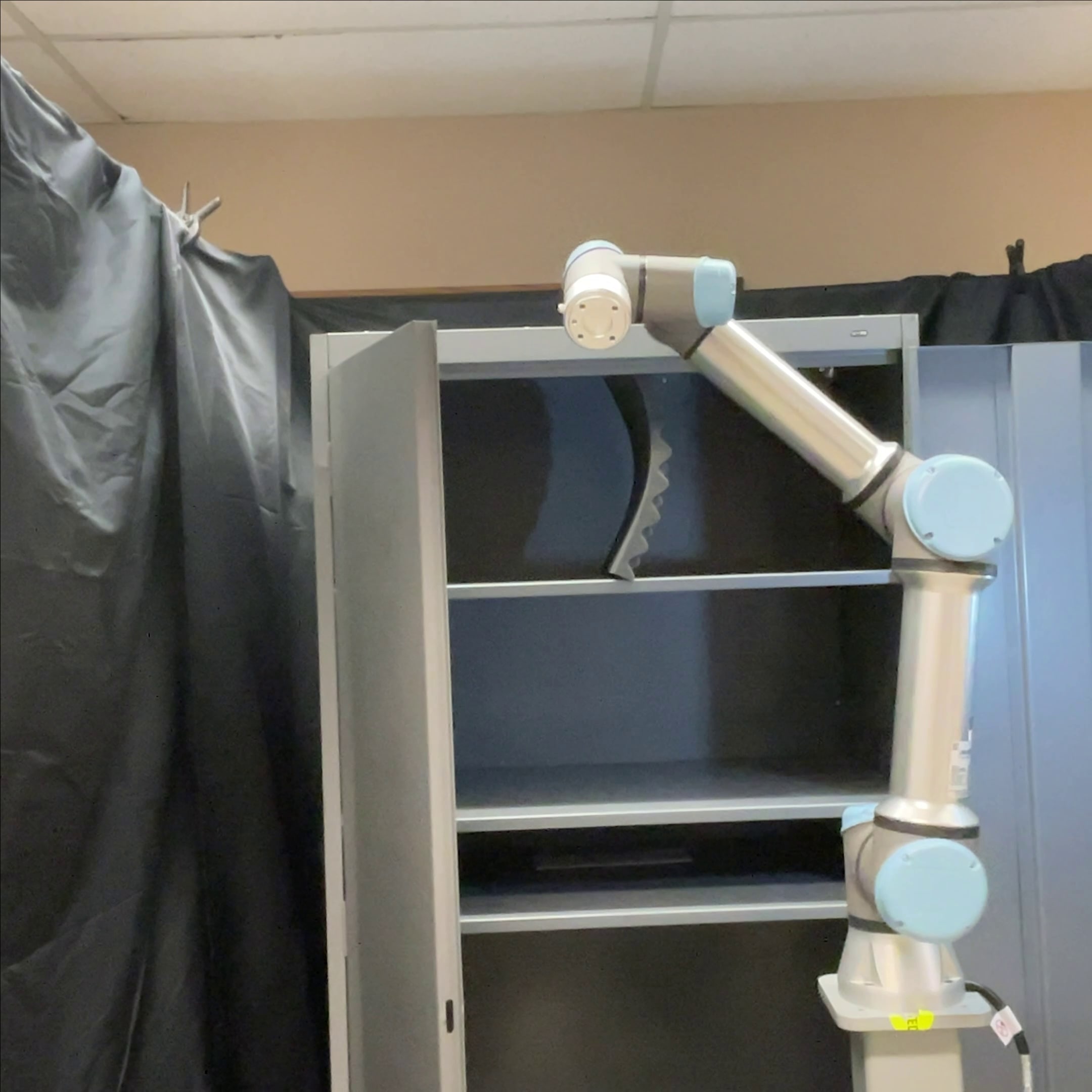}
    \includegraphics[width=0.19\textwidth]{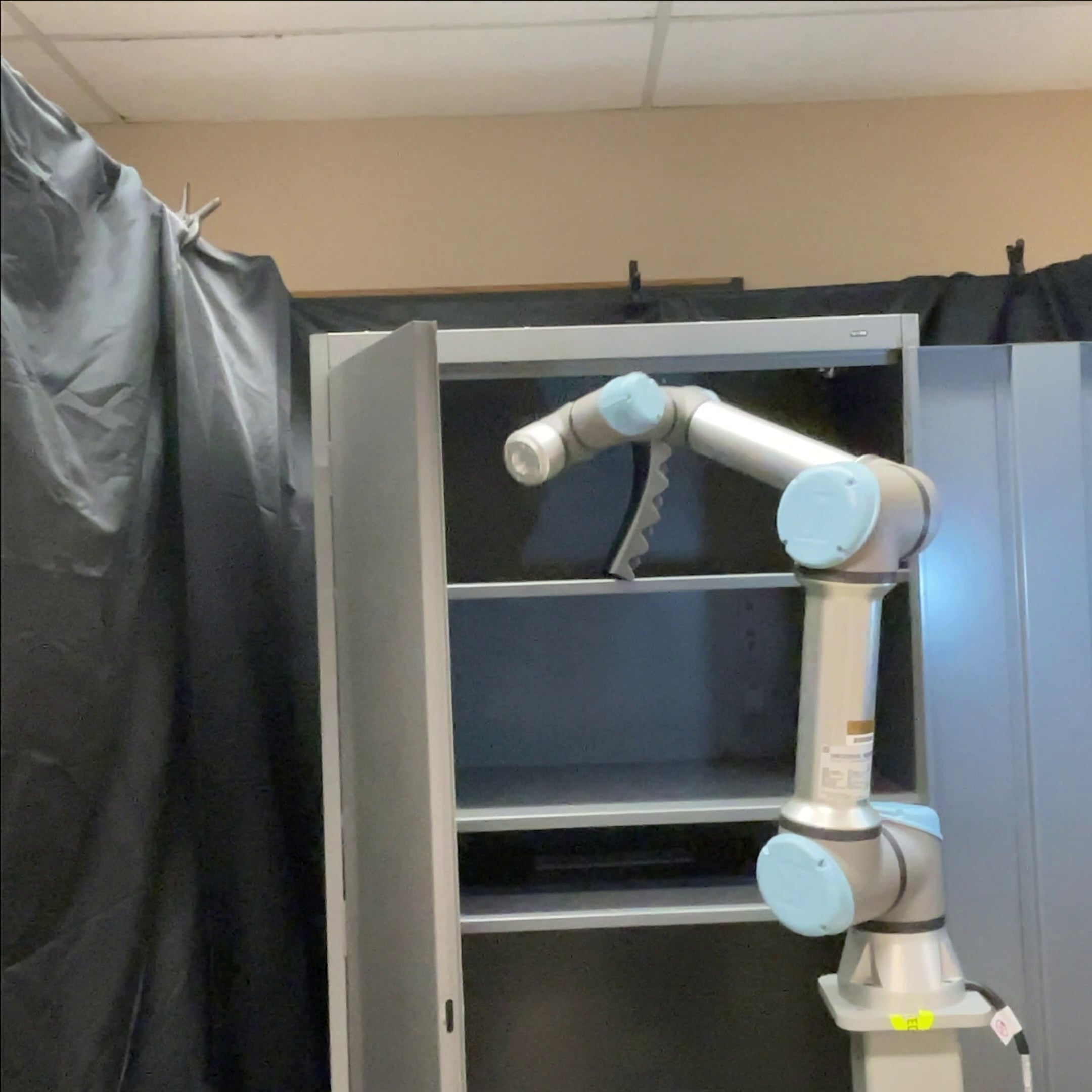}
    \includegraphics[width=0.19\textwidth]{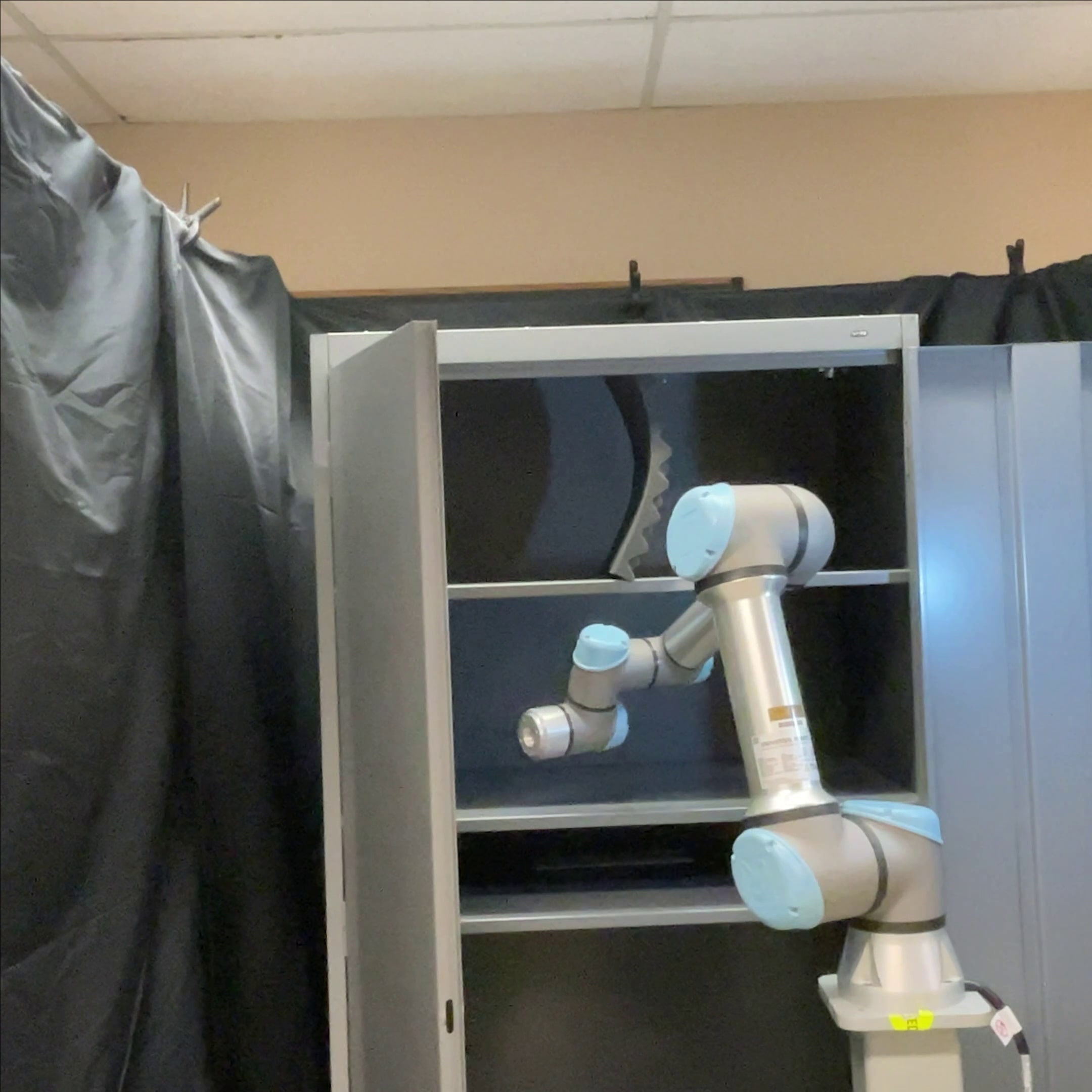} 
     \end{subfigure}
    
    \caption{ \justifying Two different real-world manipulator cases: the first case shows the manipulator crossing an obstacle at the top level to reach deeper into the narrow passage, and the second case shows the manipulator avoiding the cabinet's door to reach the given target.  }
    \label{fig:arm_real}
    \vspace{-0.2in}
\end{figure*}

    \textbf{C3D:} Our C3D environments (Fig. \ref{fig:c3d}) contain randomly placed ten boxes with different sizes in each of the eight scenarios. Furthermore, we randomly sample 1000 start and goal pairs for motion planning in each scenario to create a test dataset. Finally, we compare our method's performance with all baselines. Fig. \ref{fig:c3d} shows the paths in two example cases where our method, NTFields, IEF3D, FMM, RRT*, LazyPRM*, and RRT-Connect path solutions are illustrated in orange, red, cyan, green, brown, pink, and yellow colors, respectively. It can be seen that our method, NTFields, IEF3D, and FMM paths are smoother than RRT* and Lazy PRM* because of the obstacle margin. The obstacle margin allows safe maneuvering around the obstacles. The table in Fig. \ref{fig:c3d} presents the statistical comparison of all methods. The computational time of our method is about 2 times faster than NTFields, IEF3D, and RRT-Connect and over 30 times faster than FMM, RRT*, and LazyPRM*. Although RRT-Connect is fast for sparsely distributed obstacles, it is still slower than neural motion planners. Furthermore, all methods have near 100\% high success rate. Note that IEF3D requires expert data for training, whereas NTFields requires the training of eight separate neural models for the eight C3D scenarios. Hence NTFields add a significant computational load at training times. In contrast, our approach generalizes to eight C3D scenarios through one-time training of a single neural model and requires no expert data.

\textbf{Gibson:} Our Gibson environments (Fig. \ref{fig:gibson}) demonstrate home-like complex scenarios. We randomly sample 500 start and goal pairs for motion planning in each environment. We compare our method's performance with NTFields, IEF3D, FMM, RRT*, LazyPRM*, and RRT-Connect. Fig. \ref{fig:gibson} shows the paths where our method, NTFields, IEF3D, FMM, LazyPRM*, and RRT-Connect path solutions are illustrated in orange, red, cyan, green, pink, and yellow colors, respectively. We exclude RRT* for the cases presented in the figure as it could not find a valid within 10 seconds limit. Our method, NTFields, IEF3D, and FMM results show similar smooth paths because of the obstacle safety margin, whereas, RRT-Connect and LazyPRM* have shorter path lengths due to a smaller safety margin. The table in Fig. \ref{fig:gibson} presents the statistical results. Our method's computational time is about 2 times faster than NTFields and IEF3D and over 30 times faster than FMM, RRT*, LazyPRM*, and RRT-Connect. Furthermore, our method and FMM achieve about 98\% high success rate, while NTFields, IEF3D, RRT*, and LazyPRM* achieve about 92\% success rate. Although RRT-Connect exhibits the highest success rate, it is still 40 times slower than our method. Finally, our results validate that our progressive learning approach enables physics-informed NMPs to achieve higher performance without needing any demonstration trajectories for learning and outperform prior methods.

\begin{figure}[t]
\centering
\begin{subfigure}[b]{0.7\textwidth}
    
        \includegraphics[width=0.35\textwidth,trim=21.5cm 1.3cm 15.6cm 1.2cm,clip]{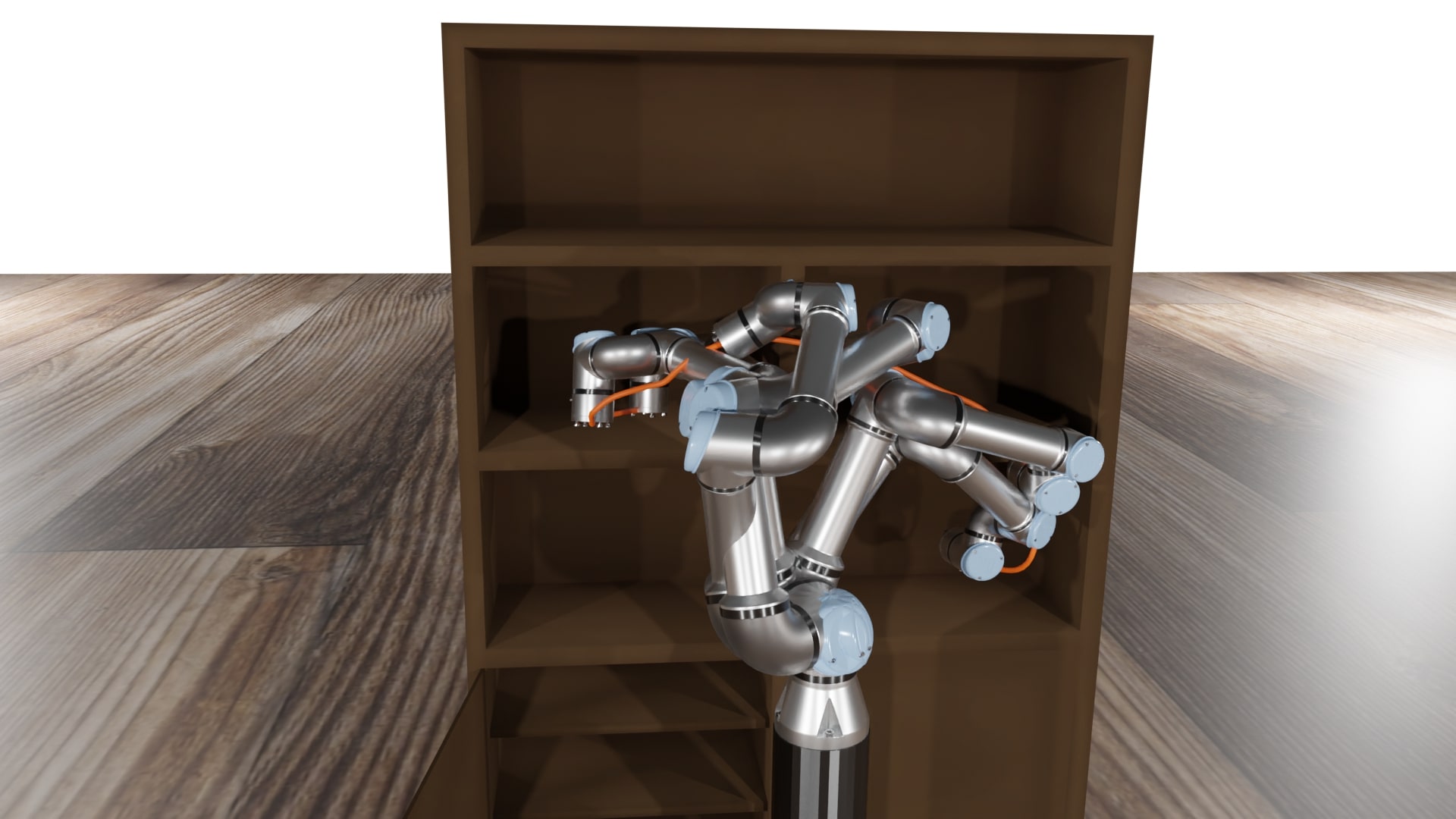}
        \includegraphics[width=0.35\textwidth,trim=21.5cm 1.3cm 15.6cm 1.2cm,clip]{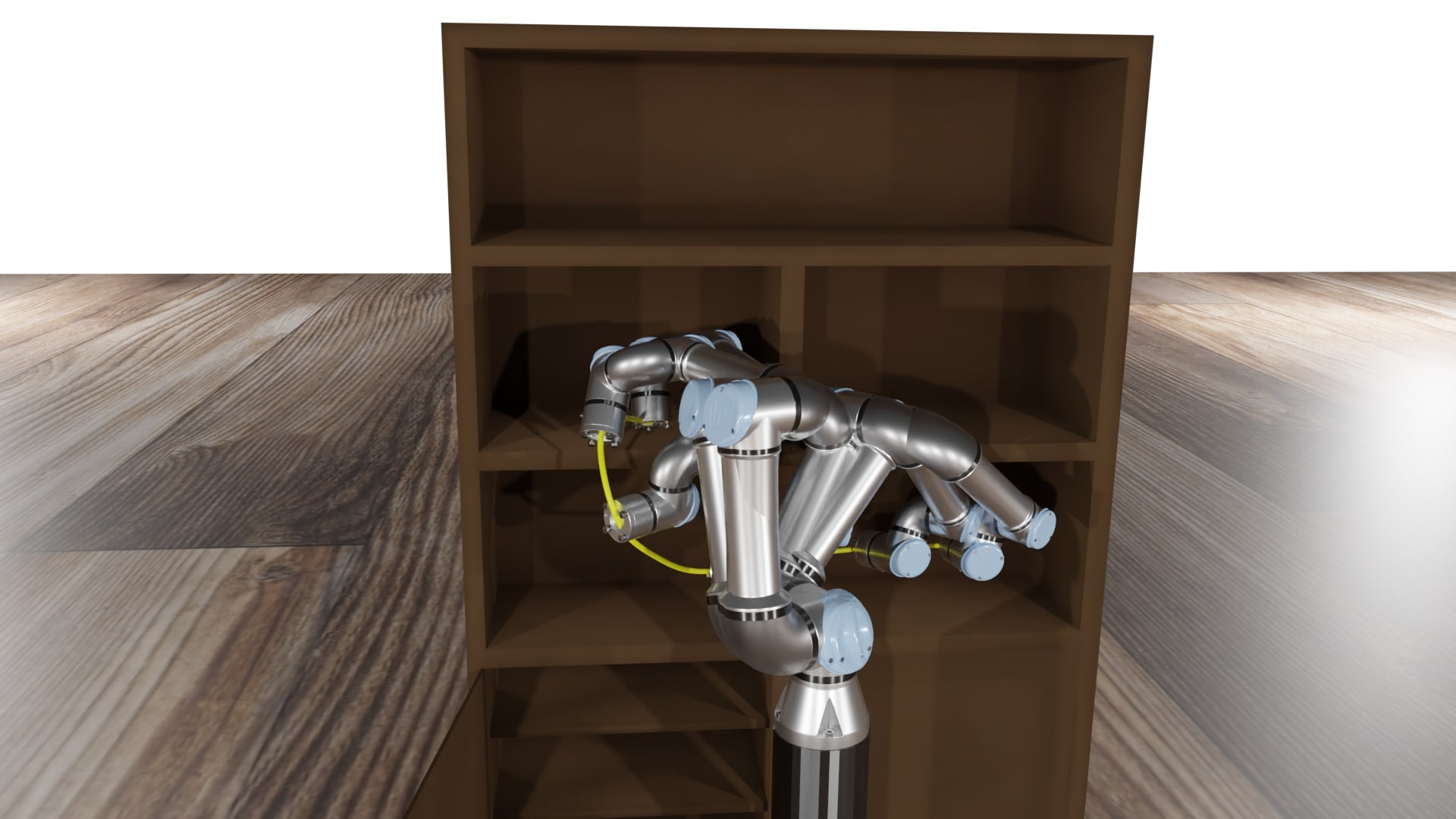}
    
    
    \label{fig:arm_sim}
    \vspace{0.1in}
\end{subfigure}

\begin{subtable}[b]{0.5\textwidth}
\centering
\footnotesize
\setlength{\tabcolsep}{5pt}
\begin{tabular}{ccccc}
\hline
Manipulator & time (sec)   &  length & {safe margin} & sr(\%)\\ 
\hline
Ours  & $0.03 \pm 0.00$ & $0.43 \pm 0.10$ & {$0.04 \pm 0.00$} & 92.0 \\
       NTFields  & $0.05 \pm 0.00$ & $0.38 \pm 0.06$ & {$0.04 \pm 0.00$} & 84.5 \\
       {RRT*}   & $5.16 \pm 0.01$  & $0.52 \pm 0.36$ & $0.04 \pm 0.00$ & 67.0\\
       { LazyPRM*}   & {$2.79 \pm 0.48$}  & {$0.76 \pm 0.80$} & {$0.04 \pm 0.00$} & 86.0\\
{RRT-Connect}   & $1.08 \pm 0.69$  & $1.14 \pm 0.23$ & $0.02 \pm 0.00$ & 87.5\\
\hline
\end{tabular}
\label{arm}
\end{subtable}
\caption{\justifying Our method (left) and RRT-Connect (right) in a challenging case in the simulated environment: the manipulator crosses two relatively thin obstacles to move from the middle (start) to the bottom (goal) shelf. The table shows statistical results on 2$\times$100 different starts and goals for simulated and real-world manipulator environments.}
\label{fig:arm_sim}
\vspace{-0.25in}
\end{figure}
\textbf{6-DOF Manipulator}

This section shows our method for 6-DOF UR5e robots in simulated and real-world environments. We chose a cabinet with narrow passages for the environment. In the simulation, we directly load a cabinet mesh, whereas, for real setup, we use Dot3D with RealSense camera to scan and create a point cloud of an actual cabinet. To form our test set, we randomly sampled 2$\times$100 start and goal configuration pairs for simulated and real-world environments. 

The table in Fig. \ref{fig:arm_sim} compares our method, NTField, RRT*, Lazy-PRM*, and RRT-Connect in both scenarios. We exclude IEF3D due to large data generation and training times. In the table, it can be seen that our method achieves the highest success rate with the shortest execution time, demonstrating the effectiveness of our progressive learning approach in complex, narrow passage environments.

 Fig. \ref{fig:arm_sim} shows the execution of our method (left) and RRT-Connect (right) in a challenging case in the simulated environment and the table underneath presents the overall statistical comparison of the indicated methods on the testing dataset. In the presented scenario, the UR5e robot's end effector starts from the middle shelf of the cabinet and crosses two relatively thin obstacles to the bottom shelf of the cabinet without collision. In this particular situation, NTField could not find a solution whereas our method took 0.07 seconds to get a 0.83 length path with a safe margin of 0.03, and RRT-Connect took 20.13 seconds to get a 0.90 length path with a safe margin of 0.02. For real-world experiments, in Fig. \ref{fig:real_arm1}, we show a challenging path that the robot went from the initial pose to make its end effect go deep into the cabinet. Finally, we demonstrate additional real-world experiments in Fig. \ref{fig:arm_real}, which depicts two cases: the first shows the manipulator crossing an obstacle at the top level, and the second shows the manipulator avoiding the cabinet's door to reach the given target. 
\begin{table}[b]
\centering
\begin{subtable}[b]{0.5\textwidth}
\centering
\footnotesize
\setlength{\tabcolsep}{5pt}
\begin{tabular}{cccc}
\hline
 Generation Time & C3D   &  Gibson & Manipulator \\ 
\hline
Ours  & $1s $ & $2s $ & {$56s $}  \\
       NTFields  & $1s $ & $2s $ & {$56s $}  \\
        IEF3D   & {$21h $}  & {$5h$} & -  \\
\hline
\end{tabular}
\label{traintime}
\vspace{0.1in}
\end{subtable}

\begin{subtable}[h]{0.5\textwidth}
\centering
\footnotesize
\setlength{\tabcolsep}{5pt}
\begin{tabular}{cccc}
\hline
 Training Time & C3D   &  Gibson & Manipulator \\ 
\hline
Ours  & $25h$ & $26h$ & {$26h$}  \\
       NTFields  & $57h$ & $32h$ & {$16h$}  \\
        IEF3D   & {$11h$}  & {$7h$} & - \\
\hline
\end{tabular}
\label{traintime}
\end{subtable}
\label{time}
\caption{Data generation and training times of our approach, NTFields, and IEF3D in different scenarios.}
\end{table}

\subsection{Data Generation and Training Time Analysis}
The following table presents the data generation and training times of our method, NTFields, and IEF3D. It can be seen that our data generation time is significantly low, similar to NTFields, as we only need to compute robot obstacle distance. In contrast, IEF3D takes several hours in data generation as it requires expert trajectories from a classical planner, FMM, for supervised learning. Furthermore, since IEF3D cannot generalize to high DOF scenarios, we exclude it in the 6-DOF manipulator environment. Regarding training times, NTFields take the longest time as it requires training a separate model for each environment, i.e., eight models for C3D, two for Gibson, and one for the manipulator. Our method training times are much lower than NTFields as our one model can generalize to multiple environments. However, our model's training times are slower than IEF3D, primarily because our method has to compute the Laplacian during training.

\section{Discussions, Conclusions, and Future Work} 
\label{sec:conclusion}

We propose a novel progressive learning framework to train physics-informed NMPs by solving the Eikonal equation without expert demonstration. Our method deals with the PDE-solving challenges in physics-informed NMPs such as NTFields \citep{ni2023ntfields}. First, we propose a progressive speed scheduling strategy that begins with finding a simple PDE solution at constant high speed and then gradually decreases the speed near the obstacle for finding a new solution. Second, we propose to use the viscosity term for the Eikonal equation and convert a nonlinear PDE to a semi-linear PDE, which is easy for a neural network to solve. Thus our method solves the Eikonal equation more precisely and efficiently and increases the overall performance in solving motion planning problems than prior methods. Additionally, our method requires fewer neural network parameters due to our progressive learning strategy than NTFields, leading to computationally efficient physics-informed NMPs' training and planning. Furthermore, we also demonstrate that our method scales to multiple environments and complex scenarios, such as real-world narrow-passage planning with a 6-DOF UR5e manipulator. 

Although our method can scale to multiple environments and real-world setups and outperform prior methods with expert demonstration data, a few limitations, highlighted in the following, will still be the focus of our future research directions. First, our method cannot generalize to unseen environments and only scales to given multiple scenarios. Therefore, one of our future directions will be to explore novel environment encoding strategies to make physics-informed NMP generalize to the novel, never-before-seen environments. Second, the success rate of our approach in the 6-DOF manipulator is around 92\%, which is better than previous approaches such as NTFields, but there is still a need for improvement to make it close to 100\% performance. Hence, another direction of our future work will be to investigate the Eikonal equation properties further to discover ways to overcome the challenges in solving high-dimensional robot manipulation problems. Lastly, aside from addressing a few limitations, we also aim to explore novel PDE formulations to train physics-informed NMPs to solve motion planning under dynamic and manifold constraints. 
\bibliographystyle{plainnat}
\bibliography{references}

\end{document}